\documentclass[sigconf]{acmart}

\usepackage{amsmath}
\usepackage{bbm}
\usepackage{graphicx}
\usepackage{caption}
\usepackage{mathtools}

\usepackage{url}
\usepackage{multirow}
\usepackage{float}
\usepackage[ruled, vlined,algo2e,linesnumbered]{algorithm2e}
\usepackage{tikz}
\usepackage{xcolor}
\usepackage{tkz-euclide,subfigure}
\usepackage{chemfig}
\usetikzlibrary{shapes}
\newlength{\R}

\makeatletter
\newcommand{\nosemic}{\renewcommand{\@endalgocfline}{\relax}}
\newcommand{\dosemic}{\renewcommand{\@endalgocfline}{\algocf@endline}}

\newcommand{\matrixSym}[1]{\ensuremath{\mathbf{\uppercase{#1}}}}

\newcommand\footnoteref[1]{\protected@xdef\@thefnmark{\ref{#1}}\@footnotemark}
\newcommand{\RN}[1]{%
	\textup{\uppercase\expandafter{\romannumeral#1}}%
}
\DeclareMathAlphabet{\mathcal}{OMS}{cmsy}{m}{n}

\newtheorem{theorem}{Theorem}
\newtheorem{definition}{Definition}

\usepackage{flushend}
\usepackage{balance}
\setcopyright{acmcopyright}
\copyrightyear{2020}
\acmYear{2020}
\acmDOI{10.1145/1122445.1122456}

\acmConference[Woodstock '20]{Woodstock '18: ACM Symposium on Neural
	Gaze Detection}{June 03--05, 2020}{Woodstock, NY}
\acmBooktitle{Woodstock '20: ACM Symposium on Neural Gaze Detection,
	June 03--05, 2020, Woodstock, NY}
\acmPrice{15.00}
\acmISBN{978-1-4503-XXXX-X/20/06}

\begin{document}
	
\sloppy

\citestyle{acmnumeric}

\title{Tree++: Truncated Tree Based Graph Kernels}

\author{Wei Ye$^1$, Zhen Wang$^2$, Rachel Redberg$^1$, Ambuj Singh$^1$}
\affiliation{%
	\institution{$^1$University of California, Santa Barbara}
}
\affiliation{%
	\institution{$^2$Columbia University}
}
\email{{weiye, rredberg, ambuj}@cs.ucsb.edu}
\email{zw2501@columbia.edu}

\renewcommand{\shortauthors}{We Ye et al.}

\begin{abstract}
Graph-structured data arise ubiquitously in many application domains. A fundamental problem is to quantify their similarities. Graph kernels are often used for this purpose, which decompose graphs into substructures and compare these substructures. However, most of the existing graph kernels do not have the property of scale-adaptivity, i.e., they cannot compare graphs at multiple levels of granularities. Many real-world graphs such as molecules exhibit structure at varying levels of granularities. To tackle this problem, we propose a new graph kernel called \textsc{Tree++} in this paper. At the heart of \textsc{Tree++} is a graph kernel called the path-pattern graph kernel. The path-pattern graph kernel first builds a truncated BFS tree rooted at each vertex and then uses paths from the root to every vertex in the truncated BFS tree as features to represent graphs. The path-pattern graph kernel can only capture graph similarity at fine granularities. In order to capture graph similarity at coarse granularities, we incorporate a new concept called super path into it. The super path contains truncated BFS trees rooted at the vertices in a path. Our evaluation on a variety of real-world graphs demonstrates that \textsc{Tree++} achieves the best classification accuracy compared with previous graph kernels.
\end{abstract}
	

\maketitle
\renewcommand{\shortauthors}{W. Ye et al.}

\section{Introduction}\label{intro}
Structured data are ubiquitous in many application domains. Examples include proteins or molecules in bioinformatics, communities in social networks, text documents in natural language processing, and images annotated with semantics in computer vision. Graphs are naturally used to represent these structured data. One fundamental problem with graph-structured data is to quantify their similarities which can be used for downstream tasks such as classification. For example, chemical compounds can be represented as graphs, where vertices represent atoms, edges represent chemical bonds, and vertex labels represent the types of atoms. We compute their similarities for classifying them into different classes. In the pharmaceutical industry, the molecule-based drug discovery needs to find similar molecules with increased efficacy and safety against a specific disease.

\begin{figure}[htb]
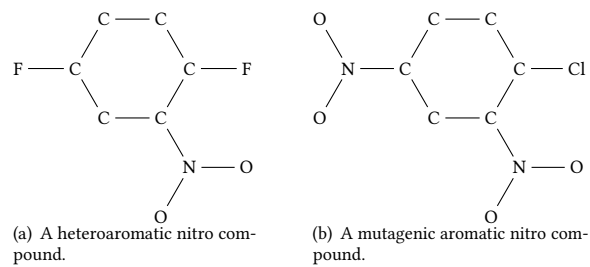
 
	\hspace*{\fill}
	\centering
	\subfigure[A heteroaromatic nitro compound.]{{\chemfig[][scale=0.8]{C(-[:180]F)*6(-C-C(-[:300]N(-[:0]O)(-[:240]O))-C(-[:0]F)-C-C-)}}}
	\hfill
	\hfill
	\centering
	\subfigure[A mutagenic aromatic nitro compound.]{{\chemfig[][scale=0.8]{C(-[:180]N(-[:120]O)(-[:240]O))*6(-C-C(-[:300]N(-[:0]O)(-[:240]O))-C(-[:0]Cl)-C-C-)}}}
	\hspace*{\fill}
	\caption{Two chemical compounds from the MUTAG \cite{kriege2012subgraph,debnath1991structure} dataset. Explicit hydrogen atoms have been removed from the original dataset. Edges represent four chemical bond types, i.e., single, double, triple or aromatic. (We do not show the edge type here for brevity.) The labels of vertices represent the types of atoms.}
	\label{fig:mutag}
\end{figure}

Figure \ref{fig:mutag} shows two chemical compounds from the MUTAG \cite{kriege2012subgraph, debnath1991structure} dataset which has 188 chemical compounds and can be divided into two classes. We can observe from Figure \ref{fig:mutag}(a) and (b) that the numbers of the atoms C (Carbon), N (Nitrogen), F (Fluorine), O (Oxygen), and Cl (Chlorine), and their varying combinations make the functions of these two chemical compounds different. We can also observe that chemical compounds (graphs) can be of arbitrary size and shape, which makes most of the machine learning methods not applicable to graphs because most of them can only handle objects of a fixed size. Tsitsulin et al. in their paper NetLSD \cite{tsitsulin2018netlsd} argued that an ideal method for graph comparison should fulfill three desiderata. The first one is permutation-invariance which means the method should be invariant to the ordering of vertices; The second one is scale-adaptivity which means the method should have different levels of granularities for comparing graphs. The last one is size-invariance which means the method can compare graphs of different sizes.

Graph kernels have been developed and widely used to measure the similarities between graph-structured data. Graph kernels are instances of the family of R-convolution kernels \cite{haussler1999convolution}. The key idea is to recursively decompose graphs into their substructures such as graphlets \cite{shervashidze2009efficient}, trees \cite{shervashidze2009fast, shervashidze2011weisfeiler}, walks \cite{vishwanathan2010graph}, paths \cite{borgwardt2005shortest}, and then compare these substructures from two graphs. A typical definition for graph kernels is $\mathcal{K}(\mathsf{G}_1, \mathsf{G}_2) = \sum_{S\in \mathcal{S}} \psi(\mathsf{G}_1, S)\psi(\mathsf{G}_2, S)$, where $\mathcal{S}$ contains all unique substructures from two graphs, and $\psi(\mathsf{G}_i, S)$ represents the number of occurrences of the unique substructure $S$ in the graph $\mathsf{G}_i, (i=1, 2)$.

\begin{figure}[htb]
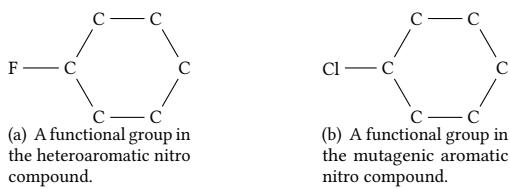
 
	\hspace*{\fill}
	\centering
	\subfigure[A functional group in the heteroaromatic nitro compound.]{{\chemfig[][scale=0.8]{C(-[:180]F)*6(-C-C-C-C-C-)}}}
	\hfill
	\hfill
	\centering
	\subfigure[A functional group in the mutagenic aromatic nitro compound.]{{\chemfig[][scale=0.8]{C(-[:180]Cl)*6(-C-C-C-C-C-)}}}
	\hspace*{\fill}
	\caption{Functional groups in the two chemical compounds from the MUTAG dataset.}
	\label{fig:functional}
\end{figure}

In the real world, many graphs such as molecules have structures at multiple levels of granularities. Graph kernels should not only capture the overall shape of graphs (whether they are more like a chain, a ring, a chain that branches, etc.), but also small structures of graphs such as chemical bonds and functional groups. For example, a graph kernel should capture that the chemical bond \chemfig{C-F} in the heteroaromatic nitro compound (Figure \ref{fig:mutag}(a)) is different from the chemical bond \chemfig{C-Cl} in the mutagenic aromatic nitro compound (Figure \ref{fig:mutag}(b)). In addition, a graph kernel should capture that the functional groups (as shown in Figure \ref{fig:functional}) in the two chemical compounds (as shown in Figure \ref{fig:mutag}) are different. Most of the existing graph kernels only have two properties, i.e., permutation-invariance and size-invariance. They cannot capture graph similarity at multiple levels of granularities. For instance, the very popular Weisfeiler-Lehman subtree kernel (WL) \cite{shervashidze2009fast, shervashidze2011weisfeiler} builds a subtree of height $h$ at each vertex and then counts the occurrences of each kind of subtree in the graph. WL can only capture the graph similarity at coarse granularities, because subtrees can only consider neighborhood structures of vertices. The shortest-path graph kernel \cite{borgwardt2005shortest} counts the number of pairs of shortest paths which have the same source and sink labels and the same length in two graphs. It can only capture the graph similarity at fine granularities, because shortest-paths do not consider neighborhood structures. The Multiscale Laplacian Graph Kernel (MLG) \cite{kondor2016multiscale} is the first graph kernel that can handle substructures at multiple levels of granularities, by building a hierarchy of nested subgraphs. However, MLG needs to invert the graph Laplacian matrix and thus its running time is very high as can be seen from Table \ref{tab:runtime} in Section \ref{result}.

In this paper, we propose a new graph kernel \textsc{Tree++} that can compare graphs at multiple levels of granularities. To this end, we first develop a base kernel called the path-pattern graph kernel that decomposes a graph into paths. For each vertex in a graph, we build a truncated BFS tree of depth $d$ rooted at it, and lists all the paths from the root to every vertex in the truncated BFS tree. Then, we compare two graphs by counting the number of occurrences of each unique path in them. We prove that the path-pattern graph kernel is positive semi-definite. The path-pattern graph kernel can only compare graphs at fine granularities. To compare graphs at coarse granularities, we extend the definition of a path in a graph and define a new concept \textit{super path}. Each element in a path is a distinct vertex while each element in a super path is a truncated BFS tree of depth $k$ rooted at the distinct vertices in a path. $k$ could be zero, and in this case, a super path degenerates to a path. Incorporated with the concept of super path, the path-pattern graph kernel can capture graph similarity at different levels of granularities, from the atomic substructure \textit{path} to the community substructure \textit{structural identity}.

Our contributions in this paper are summarized as follows: 
\begin{itemize}
	\item We propose the path-pattern graph kernel that can capture graph similarity at fine granularities.
	\item We propose a new concept of super path whose elements can be trees. After incorporating the concept of super path into the path-pattern graph kernel, it can capture graph similarity at coarse granularities.
	\item We call our final graph kernel \textsc{Tree++} as it employs truncated BFS trees for comparing graphs both at fine granularities and coarse granularities. \textsc{Tree++} runs very fast and scales up easily to graphs with thousands of vertices.
	\item \textsc{Tree++} achieves the best classification accuracy on most of the real-world datasets. 
\end{itemize} 

The paper is organized as follows: Section 2 discusses related work. Section 3 covers the core ideas and theory behind our approach, including the path-pattern graph kernel, the concept of super path, and the \textsc{Tree++} graph kernel. Using real-world datasets, Sections 4  and 5 compare \textsc{Tree++} with related techniques. Section 6 makes some discussions and Section 7 concludes the paper.

\section{Related Work}\label{relatedWork}
The first family of graph kernels is based on walks and paths, which first decompose a graph into random walks \cite{gartner2003graph, kashima2003marginalized, vishwanathan2010graph, neumann2012efficient, zhang2018retgk} or paths \cite{borgwardt2005shortest}, and then compute the number of matching pairs of them. G\"{a}rtner et al. \cite{gartner2003graph} investigate two approaches to compute graph kernels: one uses the length of all walks between each pair of vertices to define the graph similarity; the other defines one feature for every possible label sequence and then counts the number of walks in the direct product graph of two graphs matching the label sequence, of which the time complexity is $\mathcal{O}(n^6)$. If using some advanced approximation methods \cite{vishwanathan2010graph}, the time complexity could be decreased to $\mathcal{O}(n^3)$. Kashima et al. \cite{kashima2003marginalized} use random walks to generate label paths. The graph kernel is defined as the inner product of the count vector averaged over all possible label paths. Propagation kernels \cite{neumann2016propagation} leverage early-stage distributions of random walks to capture structural information hidden in vertex neighborhood. RetGK \cite{zhang2018retgk} introduces a structural role descriptor for vertices, i.e., the return probabilities features (RPF) generated by random walks. The RPF is then embedded into the Hilbert space where the corresponding graph kernels are derived. Borgwardt et al. \cite{borgwardt2005shortest} propose graph kernels based on shortest paths in graphs. It counts the number of pairs of shortest paths which have the same source and sink labels and the same length in two graphs. If the original graph is fully connected, the pairwise comparison of all edges in both graphs will cost $\mathcal{O}(n^4)$.

The second family of graph kernels is based on subgraphs, which include these kernels \cite{shervashidze2009efficient, costa2010fast, horvath2004cyclic, kondor2016multiscale} that decompose a graph into small subgraph patterns of size $k$ nodes, where $k \in \{3, 4, 5\}$. And graphs are represented as the number of all types of subgraph patterns. The subgraph patterns are called graphlets \cite{prvzulj2004modeling}. Exhaustive enumeration of all graphlets are prohibitively expensive ($\mathcal{O}(n^k)$). Thus, Shervashidze et al. \cite{shervashidze2009efficient} propose two theoretically grounded acceleration schemes. The first one uses the method of random sampling, which is motivated by the idea that the more sufficient number of random samples is drawn, the closer the empirical distribution to the actual distribution of graphlets in a graph. The second one exploits the algorithms for efficiently counting graphlets in graphs of low degree. Costa et al. \cite{costa2010fast} propose a novel graph kernel called the Neighborhood Subgraph Pairwise Distance Kernel (NSPDK) to decompose a graph into all pairs of neighborhood subgraphs of small radium at increasing distances. The authors first compute a fast graph invariant string encoding for the pairs of neighborhood subgraphs via a label function that assigns labels from a finite alphabet $\Sigma$ to the vertices in the pairs of neighborhood subgraphs. Then a hash function is used to transform the strings to natural numbers. MLG \cite{kondor2016multiscale} is developed for capturing graph structures at a range of different scales, by building a hierarchy of nested subgraphs.

The third family of graph kernels is based on subtree patterns, which decompose graphs into subtree patterns and then count the number of common subtree patterns in two graphs. Ramon et al. \cite{ramon2003expressivity} construct a graph kernel considering the subtree patterns which are rooted subgraphs at vertices. Every subtree pattern has a tree-structured signature. For each possible subtree pattern signature, the paper associates a feature of which the value is the number of times that a subtree of that signature occurs in graphs. For all pairs of vertices from two graphs, the subtree-pattern kernel counts all pairs of matching subtrees of the same signature of height less than or equal to $d$. Mah\'{e} et al. \cite{mahe2009graph} revisit and extend the subtree-pattern kernel proposed in \cite{ramon2003expressivity}  by introducing a parameter to control the complexity of the subtrees, varying from common walks to large common subtrees. Weisfeiler-Lehman subtree kernel (WL) \cite{shervashidze2009fast, shervashidze2011weisfeiler} is based on the Weisfeiler-Lehman test of isomorphism \cite{weisfeiler1968reduction} for graphs. The Weisfeiler-Lehman test of isomorphism belongs to the family of color refinement algorithms that iteratively update vertex colors (labels) until reaching the fixed number of iterations, or the vertex label sets of two graphs differ. In each iteration, the Weisfeiler-Lehman test of isomorphism algorithm augments vertex labels by concatenating their neighbors' labels and then compressing the augmented labels into new labels. The compressed labels correspond to subtree patterns. WL counts common original and compressed labels in two graphs.

Recently, some research works \cite{yanardag2015deep, kriege2016valid} focus on augmenting the existing graph kernels. DGK \cite{yanardag2015deep} deals with the problem of diagonal dominance in graph kernels. The diagonal dominance means that a graph is more similar to itself than to any other graphs in the dataset because of the sparsity of common substructures across different graphs. DGK leverages techniques from natural language processing to learn latent representations for substructures. Then the similarity matrix between substructures is computed and integrated into graph kernels. If the number of substructures is high, it will cost a lot of time and memory to compute the similarity matrix. OA \cite{kriege2016valid} develops some base kernels that generate hierarchies from which the optimal assignment kernels are computed. The optimal assignment kernels can provide a more valid notion of graph similarity. The authors finally integrate the optimal assignment kernels into the Weisfeiler-Lehman subtree kernel. In addition to the above-described literature, there are also some literature \cite{kong2011dual, tsitsulin2018netlsd, lee2018graph, nikolentzos2017matching, niepert2016learning, verma2017hunt} for graph classification that are related to our work.

The graph kernels elaborated above are only for graphs with discrete vertex labels (attributes) or no vertex labels. Recently, researchers begin to focus on the developments of graph kernels on graphs with continuous attributes. GraphHopper \cite{feragen2013scalable} is an extention of the shortest-path kernel. Instead of comparing paths based on the products of kernels on their lengths and endpoints, GraphHopper compares paths through kernels on the encountered vertices while hopping along shortest paths. The discriptor matching (DM) kernel \cite{su2016fast} maps every graph into a set of vectors (descriptors) which integrate both the attribute and neighborhood information of vertices, and then uses a set-of-vector matching kernel \cite{grauman2007approximate} to measure graph similarity. HGK \cite{morris2016faster} is a general framework to extend graph kernels from discrete attributes to continuous attributes. The main idea is to iteratively map continuous attributes to discrete labels by randomized hash functions. Then HGK compares these discrete labeled graphs by an arbitrary graph kernel such as the Weisfeiler-Lehman subtree kernel or the shortest-path kernel. GIK \cite{orsini2015graph} proposes graph invariant kernels that exploit a vertex invariant kernel (spectral coloring kernel) to combine both the similarities of vertex labels and vertex structural roles.

\section{The Model}
In this section, we introduce a new graph kernel called \textsc{Tree++}, which is based on the base kernel called the path-pattern graph kernel. The path-pattern graph kernel employs the truncated BFS (Breadth-First Search) trees rooted at each vertex of graphs. It uses the paths from the root to any other vertex in the truncated BFS trees of depth $d$ as features to represent graphs. The path-pattern graph kernel can only capture graph similarity at fine granularities. To capture graph similarity at coarse granularities, i.e., structural identities of vertices, we first propose a new concept called super path whose elements can be trees. Then, we incorporate the concept of super path into the path-pattern graph kernel.

\subsection{Notations}\label{pre}
We first give notations used in this paper to make it self-contained. In this work, we use lower-case Roman letters (e.g.\ $a,b$) to denote scalars. We denote vectors (row) by boldface lower case letters (e.g.\ $\mathbf{x}$) and denote its $i$-th element by $\mathbf{x}(i)$. Matrices are denoted by boldface upper case letters (e.g.\ $\matrixSym{X}$). We denote entries in a matrix as $\matrixSym{X}(i, j)$. We use $\mathbf{x}=[x_1,\cdots, x_n]$ to denote creating a vector by stacking scalar $x_i$ along the columns. Similarly, we use $\mathbf{X}=[\mathbf{x}_1;\ldots; \mathbf{x}_n]$ to denote creating a matrix by stacking the vector $\mathbf{x}_i$ along the rows. Consider an undirected labeled graph $\mathsf{G}=(\mathcal{V},\mathcal{E}, l)$, where $\mathcal{V}$ is a set of graph vertices with number $|\mathcal{V}|$ of vertices, $\mathcal{E}$ is a set of graph edges with number $|\mathcal{E}|$ of edges, and $l: \mathcal{V}\rightarrow \Sigma$ is a function that assigns labels from a set of positive integers $\Sigma$ to vertices. Without loss of generality, $\lvert\Sigma\rvert\leq |\mathcal{V}|$.

An edge $e$ is denoted by two vertices $uv$ that are connected to it. In graph theory \cite{harary1969graph}, a walk is defined as a sequence of vertices, e.g., $(v_1,v_2,\cdots)$, where consecutive vertices are connected by an edge. A trail is a walk that consists of all distinct edges. A path is a trail that consists of all distinct vertices and edges. A spanning tree $\mathsf{ST}$ of a graph $\mathsf{G}$ is a subgraph that includes all of the vertices of $\mathsf{G}$, with the minimum possible number of edges. We extend this definition to the truncated spanning tree. A truncated spanning tree $\mathsf{T}$ is a subtree of the spanning tree $\mathsf{ST}$, with the same root and of the depth $d$. The depth of a subtree is the maximum length of paths between the root and any other vertex in the subtree. Two undirected labeled graphs $\mathsf{G}_1=(\mathcal{V}_1,\mathcal{E}_1, l_1)$ and $\mathsf{G}_2=(\mathcal{V}_2,\mathcal{E}_2, l_2)$ are isomorphic (denoted by $\mathsf{G}_1 \simeq \mathsf{G}_2$) if there is a bijection $\varphi: \mathcal{V}_1 \rightarrow \mathcal{V}_2$, (1) such that for any two vertices $u, v \in \mathcal{V}_1$, there is an edge $uv$ if and only if there is an edge $\varphi(u)\varphi(v)$ in $\mathsf{G}_2$; (2) and such that $l_2(\varphi(v)) = l_1(v)$.

Let $\mathcal{X}$ be a non-empty set and let $\mathcal{K}: \mathcal{X} \times \mathcal{X} \rightarrow \mathbb{R}$ be a function on the set $\mathcal{X}$. Then $\mathcal{K}$ is a kernel on $\mathcal{X}$ if there is a real Hilbert space $\mathcal{H}$ and a mapping $\phi:  \mathcal{X} \rightarrow \mathcal{H}$ such that $\mathcal{K}(x, y) = \langle\phi(x), \phi(y)\rangle$ for all $x$, $y$ in $\mathcal{X}$, where $\langle\cdot, \cdot\rangle$ denotes the inner product of $\mathcal{H}$, $\phi$ is called a feature map and $\mathcal{H}$ is called a feature space. $\mathcal{K}$ is symmetric and positive-semidefinite. In the case of graphs, let $\phi(\mathsf{G})$ denote a mapping from graph to vector which contains the counts of atomic substructures in graph $\mathsf{G}$. Then, the kernel on two graphs $\mathsf{G}_1$ and $\mathsf{G}_2$ is defined as $\mathcal{K}(\mathsf{G}_1, \mathsf{G}_2) = \langle\phi(\mathsf{G}_1), \phi(\mathsf{G}_2)\rangle$. 

\subsection{The Path-Pattern Graph Kernel}
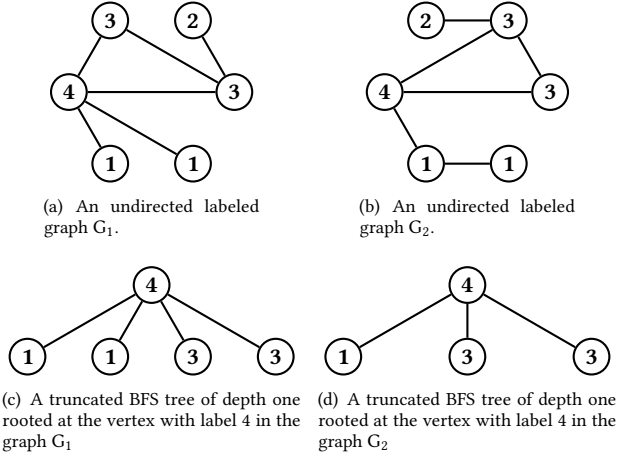
\begin{figure}[htb]
	\hspace*{\fill}
	\centering
	\subfigure[An undirected labeled graph $\mathsf{G}_1$.]{
		\begin{tikzpicture}
		[scale=1.1,every node/.style={draw, circle, thick,minimum size=1.5em,inner sep=1}]
		\node (n6) at (60: \R) {\textbf{2}};
		\node (n4) at (120: \R)  {\textbf{3}};
		\node (n5) at (180: \R)  {\textbf{4}};
		\node (n1) at (240: \R) {\textbf{1}};
		\node (n2) at (300: \R)  {\textbf{1}};
		\node (n3) at (360: \R)  {\textbf{3}};
		
		\foreach \from/\to in {n4/n5,n5/n1,n2/n5,n3/n4,n5/n3,n6/n3}
		\draw[thick] (\from) -- (\to);
		\end{tikzpicture}
	}
    \hfill
    \hfill
    \centering
     \subfigure[An undirected labeled graph $\mathsf{G}_2$.]{
    	\begin{tikzpicture}
    	[scale=1.1,every node/.style={draw, circle, thick,minimum size=1.5em,inner sep=1}]
    	\node (n6) at (60: \R) {\textbf{3}};
    	\node (n4) at (120: \R)  {\textbf{2}};
    	\node (n5) at (180: \R)  {\textbf{4}};
    	\node (n1) at (240: \R) {\textbf{1}};
    	\node (n2) at (300: \R)  {\textbf{1}};
    	\node (n3) at (360: \R)  {\textbf{3}};
    	
    	\foreach \from/\to in {n6/n4,n5/n1,n2/n1,n5/n6,n5/n3,n6/n3}
    	\draw[thick] (\from) -- (\to);
    	\end{tikzpicture}
    }
    \hspace*{\fill}
    
    \hspace*{\fill}
    \centering
   \subfigure[A truncated BFS tree of depth one rooted at the vertex with label 4 in the graph $\mathsf{G}_1$]{
   	\begin{tikzpicture}
   	[scale=1.1,level distance=8.66mm,
   	every node/.style={draw, circle, thick,minimum size=1.5em,inner sep=1},
   	level 1/.style={sibling distance=10mm,thick},
   	level 2/.style={sibling distance=10mm,thick}]
   	\node {\textbf{4}}
   	child {node {\textbf{1}}
   		child[missing]
   	}
   	child {node {\textbf{1}}
   		child[missing]
   	}
   	child {node {\textbf{3}}
   		child[missing]
   	}
   	child {node {\textbf{3}}
   		child[missing]
   	};
   	\end{tikzpicture}
   }
    \hfill
    \hfill
    \centering
    \subfigure[A truncated BFS tree of depth one rooted at the vertex with label 4 in the graph $\mathsf{G}_2$]{
    	\begin{tikzpicture}
    	[scale=1.1,level distance=8.66mm,
    	every node/.style={draw, circle, thick,minimum size=1.5em,inner sep=1},
    	level 1/.style={sibling distance=15mm,thick},
    	level 2/.style={sibling distance=15mm,thick}]
    	\node {\textbf{4}}
    	child {node {\textbf{1}}
    		child[missing]
    	}
    	child {node {\textbf{3}}
    		child[missing]
    	}
    	child {node {\textbf{3}}
    		child[missing]
    	};
    	\end{tikzpicture}
    }
    \hspace*{\fill}
	\caption{Illustration of the path patterns in graphs. $\Sigma=\{1,2,3,4\}$.}
\label{fig:example}
\end{figure}

We first define the path pattern as follows:
\begin{definition}[Path Pattern]
	Given an undirected labeled graph $\mathsf{G}=(\mathcal{V},\mathcal{E}, l)$, we build a truncated BFS tree $\mathsf{T}=(\mathcal{V}',\mathcal{E}', l)$ ($\mathcal{V}'\subseteq\mathcal{V}$ and $\mathcal{E}'\subseteq\mathcal{E}$) of depth $d$ rooted at each vertex $v \in\mathcal{V}$. The vertex $v$ is called root. For each vertex $v' \in \mathcal{V}'$, there is a path $P=\left( v,v_1,v_2,\cdots,v'\right)$ from the root $v$ to $v'$ consisting of distinct vertices and edges. The concatenated labels $l(P)=\left( l(v),l(v_1),l(v_2),\cdots,l(v')\right) $ is called a path pattern of $\mathsf{G}$.
\end{definition}
Note from this definition that a path pattern could only contain the root vertex. Figure \ref{fig:example}(a) and (b) show two example undirected labeled graph $\mathsf{G}_1$ and $\mathsf{G}_2$. Figure \ref{fig:example}(c) and (d) show two truncated BFS trees of depth $d=1$ on the vertices with label 4 in $\mathsf{G}_1$ and $\mathsf{G}_2$, respectively. To build unique BFS trees, the child vertices of each parent vertex in the BFS tree are sorted in ascending order according to their label values. If two vertices have the same label values, we sort them again in ascending order by their eigenvector centrality \cite{bonacich1987power} values. We use eigenvector centrality to measure the importance of a vertex. A vertex has high eigenvector centrality value if it is linked to by other vertices that also have high eigenvector centrality values, without implying that this vertex is highly linked. All of the path patterns of the root of the BFS tree in Figure \ref{fig:example}(c) are as follows: $(4), (4,1), (4,1), (4,3), (4,3)$. All of the path patterns of the root of the BFS tree in Figure \ref{fig:example}(d) are as follows: $(4), (4,1), (4,3), (4,3)$. On each vertex in the graph $\mathsf{G}_1$ and $\mathsf{G}_2$, we first build a truncated BFS tree of depth $d$, and then generate all of its corresponding path patterns. The multiset\footnote{ A set that can contain the same element multiple times.} $\mathcal{M}$ of the graph $\mathsf{G}$ is a set that contains all the path patterns extracted from BFS trees of depth $d$ rooted at each vertex of the graph.

Let  $\mathcal{M}_1$ and $\mathcal{M}_2$ be two multisets corresponding to the two graphs $\mathsf{G}_1$ and $\mathsf{G}_2$. Let the union of $\mathcal{M}_1$ and $\mathcal{M}_2$ be $\mathcal{U}=\mathcal{M}_1\cup\mathcal{M}_2=\{l(P_1), l(P_2),\cdots,l(P_{|\mathcal{U}|})\}$. Define a map $\psi: \{\mathsf{G}_1,\mathsf{G}_2\} \times \Sigma \rightarrow \mathbb{N}$ such that $\psi(\mathsf{G},l(P_i))$ is the number of occurrences of the path pattern $l(P_i)$ in the graph $\mathsf{G}$. The definition of the path-pattern graph kernel is given as follows:
\begin{equation}
\label{equ:kpp}
\mathcal{K}_{pp}(\mathsf{G}_1, \mathsf{G}_2) = \sum_{l(P_i)\in\mathcal{U}}\psi\left(\mathsf{G}_1, l(P_i)\right) \psi\left( \mathsf{G}_2, l(P_i)\right) 
\end{equation}

\begin{theorem}
	The path-pattern graph kernel $\mathcal{K}_{pp}$ is positive semi-definite.
\end{theorem}

\begin{proof}
The path-pattern graph kernel $\mathcal{K}_{pp}$ in Equation (\ref{equ:kpp}) can also be written as follows:
\begin{displaymath}
\mathcal{K}_{pp}(\mathsf{G}_1, \mathsf{G}_2) = \left\langle \phi(\mathsf{G}_1), \phi(\mathsf{G}_2)\right\rangle 
\end{displaymath}
where $\phi(\mathsf{G}_1)=\left[ \psi(\mathsf{G}_1,l(P_1)), \psi(\mathsf{G}_1,l(P_2)),\cdots,\psi(\mathsf{G}_1,l(P_{|\mathcal{U}|}))\right] $ and $\phi(\mathsf{G}_2)=\left[ \psi(\mathsf{G}_2,l(P_1)), \psi(\mathsf{G}_2,l(P_2)),\cdots,\psi(\mathsf{G}_2,l(P_{|\mathcal{U}|}))\right] $.
	
Inspired by earlier works on graph kernels, we can readily verify that for any vector $\mathbf{x}\in\mathbb{R}^n$, we have
\begin{equation}
\label{equ:semidefinite}
\begin{aligned}
\mathbf{x}^\intercal \mathcal{K}_{pp}\mathbf{x}&=\sum_{i,j=1}^nx_ix_j\mathcal{K}_{pp}(\mathsf{G}_i, \mathsf{G}_j)\\
&=\sum_{i,j=1}^nx_ix_j\left\langle \phi(\mathsf{G}_i), \phi(\mathsf{G}_j)\right\rangle \\
&=\left\langle \sum_{i=1}^nx_i\phi(\mathsf{G}_i),\sum_{j=1}^nx_j\phi(\mathsf{G}_j)\right\rangle\\
&=\left\| \sum_{i=1}^nx_i\phi(\mathsf{G}_i)\right\| \geq0
\end{aligned}
\end{equation}
\end{proof}

For example, if the depth of BFS tree is set to one, the multisets $\mathcal{M}_1$ and $\mathcal{M}_2$ are as follows:
\begin{displaymath}
\begin{aligned}
\mathcal{M}_1 &= \left\lbrace (1), (1,4), (1), (1,4), (4), (4,1), (4,1), (4,3), (4,3),\right. \\
&\quad \left. (3), (3,3), (3,4), (2), (2,3), (3), (3,2), (3,3), (3,4)\right\rbrace \\
\mathcal{M}_2 &= \left\lbrace (1), (1,1), (1), (1,1), (1,4), (4), (4,1), (4,3), (4,3),\right. \\
&\quad \left. (2), (2,3), (3), (3,2), (3,3), (3,4), (3), (3,3), (3,4)\right\rbrace \\
\end{aligned}
\end{displaymath}

The union of $\mathcal{M}_1$ and $\mathcal{M}_2$ is $\mathcal{U}=\mathcal{M}_1\cup\mathcal{M}_2$ which is a normal set containing unique elements. The elements are sorted lexicographically.
\begin{displaymath}
\begin{aligned}
\mathcal{U} &= \left\lbrace (1), (1,1), (1,4), (2), (2,3),\right. \\
&\quad \left. (3), (3,2), (3,3), (3,4), (4), (4,1), (4,3)\right\rbrace
\end{aligned}
\end{displaymath}

\noindent Considering that a path $uv$ is equivalent to its reversed one $vu$ in undirected graphs, we remove the repetitive path patterns in $\mathcal{U}$ and finally we have:
\begin{displaymath}
\mathcal{U} = \left\lbrace (1), (1,1), (1,4), (2), (2,3), (3), (3,3), (3,4), (4)\right\rbrace
\end{displaymath}

\noindent For each path pattern in the set $\mathcal{U}$, we count its occurrences in $\mathsf{G}_1$ and $\mathsf{G}_2$ and have the following:
\begin{displaymath}
\begin{aligned}
\phi(\mathsf{G}_1)&=\left[ \psi\left(\mathsf{G}_1, (1)\right), \psi\left(\mathsf{G}_1, (1,1)\right), \cdots, \psi\left(\mathsf{G}_1, (4)\right)\right]  \\
&=\left[  2, 0, 4, 1, 2, 2, 2, 4, 1\right]  \\
\phi(\mathsf{G}_2)&=\left[ \psi\left(\mathsf{G}_2, (1)\right), \psi\left(\mathsf{G}_2, (1,1)\right), \cdots, \psi\left(\mathsf{G}_2, (4)\right)\right]  \\
&=\left[ 2, 2, 2, 1, 2, 2, 2, 4, 1\right]  \\
\end{aligned}
\end{displaymath}

\noindent Thus $\mathcal{K}_{pp}(\mathsf{G}_1, \mathsf{G}_2)=\left\langle \phi(\mathsf{G}_1), \phi(\mathsf{G}_2)\right\rangle=42$ 

The path-pattern graph kernel will be used as the base kernel for our final \textsc{Tree++} graph kernel. We can see that the path-pattern graph kernel decomposes a graph into its substructures, i.e., paths. However,  paths cannot reveal the structural or topological information of vertices. Thus, the path-pattern graph kernel can only capture graph similarity at fine granularities. Likewise, most of the graph kernels that belong to the family of R-convolution framework \cite{haussler1999convolution} face the same problem colloquially stated as losing sight of the forest for the trees. To capture graph similarity at coarse granularities, we need to zoom out our perspectives on graphs and focus on the structural identities.

\subsection{Incorporating Structural Identity}
Structural identity is a concept to define the class of vertices in a graph by considering the graph structure and their relations to other vertices. In graphs, vertices are often associated with some functions that determine their roles in the graph. For example, each of the proteins in a protein-protein interaction (PPI) network has a specific function, such as enzyme, antibody, messenger, transport/storage, and structural component. Although such functions may also depend on the vertex and edge attributes, in this paper, we only consider their relations to the graph structures. Explicitly considering the structural identities of vertices in graphs for the design of graph kernels has been missing from the literature except the Weisfeiler–Lehman subtree kernel \cite{shervashidze2009fast, shervashidze2011weisfeiler}, Propagation kernel \cite{neumann2016propagation}, MLG \cite{kondor2016multiscale}, and RetGK \cite{zhang2018retgk}.

To incorporate the structural identity information into graph kernels, in this paper, we extend the definition of path in graphs and define super path as follows:
\begin{definition}[Super Path]
	Given an undirected labeled graph $\mathsf{G}=(\mathcal{V},\mathcal{E}, l)$, we build a truncated BFS tree $\mathsf{T}=(\mathcal{V}',\mathcal{E}', l)$ ($\mathcal{V}'\subseteq\mathcal{V}$ and $\mathcal{E}'\subseteq\mathcal{E}$) of depth $d$ rooted at each vertex $v \in\mathcal{V}$. The vertex $v$ is called root. For each vertex $v' \in \mathcal{V}'$, there is a path $P=\left( v,v_1,v_2,\cdots,v'\right)$ from the root $v$ to $v'$ consisting of distinct vertices and edges. For each vertex in $P$, we build a truncated BFS tree of depth $k$ rooted at it. The sequence of trees $S=\left(\mathsf{T}_v,\mathsf{T}_{v_1},\mathsf{T}_{v_2},\cdots,\mathsf{T}_{v'}\right)$ is called a super path.
\end{definition}

We can see that the definition of super path includes the definition of path in graphs. Path is a special case of super path when the truncated BFS tree on each distinct vertex in a path is of depth 0.

The problem now is that what is the path pattern corresponding to the super path? In other words, what is the label of each truncated BFS tree in the super path? In this case, we also need to extend the definition of the label function $l$ described in Section \ref{pre} as follows: $l: \mathsf{T}\rightarrow \Sigma$ ($\Sigma$ here is different from above. We abuse the notation.) is a function that assigns labels from a set of positive integers $\Sigma$ to trees. Thus, the definition of the path pattern for super paths is: the concatenated labels $l(S)=\left( l(\mathsf{T}_v),l(\mathsf{T}_{v_1}),l(\mathsf{T}_{v_2}),\cdots, l(\mathsf{T}_v')\right) $ is called a path pattern.

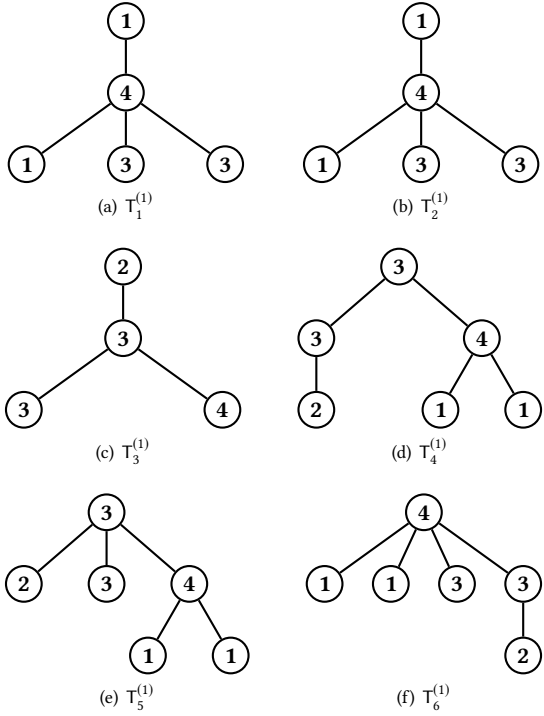
\begin{figure}[htb]
	\hspace*{\fill}
	\centering
	\subfigure[$\mathsf{T}_1^{(1)}$]{
		\begin{tikzpicture}
		[scale=1.1,level distance=8.66mm,
		every node/.style={draw, circle, thick,minimum size=1.5em,inner sep=1},
		level 1/.style={sibling distance=10mm,thick},
		level 2/.style={sibling distance=12mm,thick}]
		\node {\textbf{1}}
		child {node {\textbf{4}}
			child {node {\textbf{1}}
				child[missing]
			}
			child {node {\textbf{3}}
				child[missing]
			}
			child {node {\textbf{3}}
				child[missing]
			}
		};
		\end{tikzpicture}
	}
	\hfill
	\centering
	\subfigure[$\mathsf{T}_2^{(1)}$]{
		\begin{tikzpicture}
		[scale=1.1,level distance=8.66mm,
		every node/.style={draw, circle, thick,minimum size=1.5em,inner sep=1},
		level 1/.style={sibling distance=10mm,thick},
		level 2/.style={sibling distance=12mm,thick}]
		\node {\textbf{1}}
		child {node {\textbf{4}}
			child {node {\textbf{1}}
				child[missing]
			}
			child {node {\textbf{3}}
				child[missing]
			}
			child {node {\textbf{3}}
				child[missing]
			}
		};
		\end{tikzpicture}
	}
	\hspace*{\fill}

	\hspace*{\fill}
	\centering
	\subfigure[$\mathsf{T}_3^{(1)}$]{
		\begin{tikzpicture}
		[scale=1.1,level distance=8.66mm,
		every node/.style={draw, circle, thick,minimum size=1.5em,inner sep=1},
		level 1/.style={sibling distance=10mm,thick},
		level 2/.style={sibling distance=24mm,thick}]
		\node {\textbf{2}}
		child {node {\textbf{3}}
			child {node {\textbf{3}}
				child[missing]
			}
			child {node {\textbf{4}}
				child[missing]
			}
		};
		\end{tikzpicture}
	}
	\hfill
	\centering
	\subfigure[$\mathsf{T}_4^{(1)}$]{
		\begin{tikzpicture}
		[scale=1.1,level distance=8.66mm,
		every node/.style={draw, circle, thick,minimum size=1.5em,inner sep=1},
		level 1/.style={sibling distance=20mm,thick},
		level 2/.style={sibling distance=10mm,thick}]
		\node {\textbf{3}}
		child {node {\textbf{3}}
			child {node {\textbf{2}}
				child[missing]
			}
		}
		child {node {\textbf{4}}
			child {node {\textbf{1}}
				child[missing]
			}
			child {node {\textbf{1}}
				child[missing]
			}
		};
		\end{tikzpicture}
	}
	\hspace*{\fill}

	\hspace*{\fill}
	\centering
	\subfigure[$\mathsf{T}_5^{(1)}$]{
		\begin{tikzpicture}
		[scale=1.1,level distance=8.66mm,
		every node/.style={draw, circle, thick,minimum size=1.5em,inner sep=1},
		level 1/.style={sibling distance=10mm,thick},
		level 2/.style={sibling distance=10mm,thick}]
		\node {\textbf{3}}
		child {node {\textbf{2}}
			child[missing]
		}
		child {node {\textbf{3}}
			child[missing]
		}
		child {node {\textbf{4}}
			child {node {\textbf{1}}}
			child {node {\textbf{1}}}
		};
		\end{tikzpicture}
	}
	\hfill
	\centering
	\subfigure[$\mathsf{T}_6^{(1)}$]{
		\begin{tikzpicture}
		[scale=1.1,level distance=8.66mm,
		every node/.style={draw, circle, thick,minimum size=1.5em,inner sep=1},
		level 1/.style={sibling distance=8mm,thick},
		level 2/.style={sibling distance=10mm,thick}]
		\node {\textbf{4}}
		child {node {\textbf{1}}
			child[missing]
		}
		child {node {\textbf{1}}
			child[missing]
		}
		child {node {\textbf{3}}
			child[missing]
		}
	    child {node {\textbf{3}}
		    child {node {\textbf{2}}}
	    };
		\end{tikzpicture}
	}
	\hspace*{\fill}
	\caption{A truncated BFS tree of depth two rooted at each vertex in the undirected labeled graph $\mathsf{G}_1$.}
	\label{fig:truncatedtree1}
\end{figure}

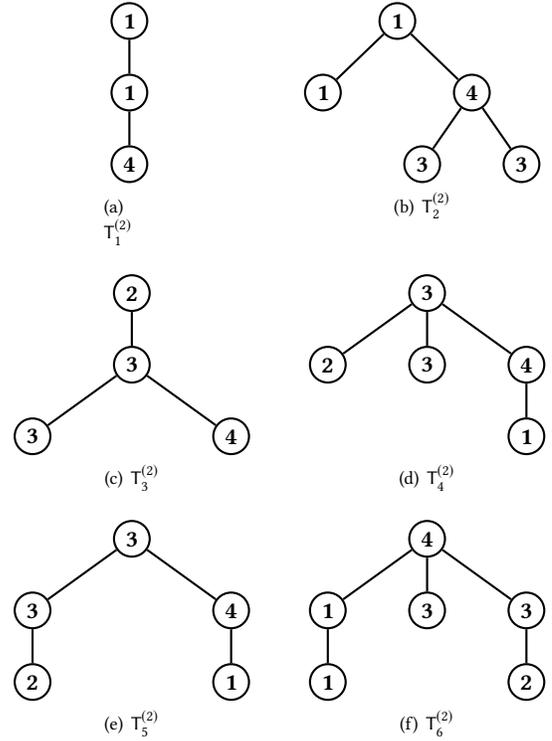
\begin{figure}[htb]
	\hspace*{\fill}
	\hfill
	\hfill
	\centering
	\subfigure[$\mathsf{T}_1^{(2)}$]{
		\begin{tikzpicture}
		[scale=1.1,level distance=8.66mm,
		every node/.style={draw, circle, thick,minimum size=1.5em,inner sep=1},
		level 1/.style={sibling distance=20mm,thick},
		level 2/.style={sibling distance=20mm,thick}]
		\node {\textbf{1}}
		child {node {\textbf{1}}
			child {node {\textbf{4}}
				child[missing]
			}
		};
		\end{tikzpicture}
	}
	\hfill
	\hfill
	\hfill
	\centering
	\subfigure[$\mathsf{T}_2^{(2)}$]{
		\begin{tikzpicture}
		[scale=1.1,level distance=8.66mm,
		every node/.style={draw, circle, thick,minimum size=1.5em,inner sep=1},
		level 1/.style={sibling distance=18mm,thick},
		level 2/.style={sibling distance=12mm,thick}]
		\node {\textbf{1}}
		child {node {\textbf{1}}
			child[missing]
		}
		child{node{\textbf{4}}
			child {node {\textbf{3}}
				child[missing]
			}
			child {node {\textbf{3}}
				child[missing]
			}
		};
		\end{tikzpicture}
	}
	\hspace*{\fill}

	\hspace*{\fill}
	\centering
	\subfigure[$\mathsf{T}_3^{(2)}$]{
		\begin{tikzpicture}
		[scale=1.1,level distance=8.66mm,
		every node/.style={draw, circle, thick,minimum size=1.5em,inner sep=1},
		level 1/.style={sibling distance=10mm,thick},
		level 2/.style={sibling distance=24mm,thick}]
		\node {\textbf{2}}
		child {node {\textbf{3}}
			child {node {\textbf{3}}
				child[missing]
			}
			child {node {\textbf{4}}
				child[missing]
			}
		};
		\end{tikzpicture}
	}
	\hfill
	\centering
	\subfigure[$\mathsf{T}_4^{(2)}$]{
		\begin{tikzpicture}
		[scale=1.1,level distance=8.66mm,
		every node/.style={draw, circle, thick,minimum size=1.5em,inner sep=1},
		level 1/.style={sibling distance=12mm,thick},
		level 2/.style={sibling distance=10mm,thick}]
		\node {\textbf{3}}
		child {node {\textbf{2}}
			child[missing]
		}
		child {node {\textbf{3}}
			child[missing]
		}
		child {node {\textbf{4}}
			child {node {\textbf{1}}
				child[missing]
			}
		};
		\end{tikzpicture}
	}
	\hspace*{\fill}
	
	\hspace*{\fill}
	\centering
	\subfigure[$\mathsf{T}_5^{(2)}$]{
		\begin{tikzpicture}
		[scale=1.1,level distance=8.66mm,
		every node/.style={draw, circle, thick,minimum size=1.5em,inner sep=1},
		level 1/.style={sibling distance=24mm,thick},
		level 2/.style={sibling distance=10mm,thick}]
		\node {\textbf{3}}
		child {node {\textbf{3}}
			child {node {\textbf{2}}}
		}
		child {node {\textbf{4}}
			child {node {\textbf{1}}}
		};
		\end{tikzpicture}
	}
	\hfill
	\centering
	\subfigure[$\mathsf{T}_6^{(2)}$]{
		\begin{tikzpicture}
		[scale=1.1,level distance=8.66mm,
		every node/.style={draw, circle, thick,minimum size=1.5em,inner sep=1},
		level 1/.style={sibling distance=12mm,thick},
		level 2/.style={sibling distance=24mm,thick}]
		\node {\textbf{4}}
		child {node {\textbf{1}}
			child {node {\textbf{1}}
				child[missing]
			}
		}
		child {node {\textbf{3}}
			child[missing]
		}
		child {node {\textbf{3}}
			child {node {\textbf{2}}
				child[missing]
			}
		};
		\end{tikzpicture}
	}
	\hspace*{\fill}
	\caption{A truncated BFS tree of depth two rooted at each vertex in the undirected labeled graph $\mathsf{G}_2$.}
	\label{fig:truncatedtree2}
\end{figure}

\begin{figure}[htb]
	\hspace*{\fill}
	\centering
	\subfigure[Relabeled $\mathsf{G}_1$.]{
		\begin{tikzpicture}
		[scale=1.1,every node/.style={draw, circle, thick,minimum size=1.5em,inner sep=1}]
		\node (n6) at (60: \R) {\textbf{4}};
		\node (n4) at (120: \R)  {\textbf{8}};
		\node (n5) at (180: \R)  {\textbf{9}};
		\node (n1) at (240: \R) {\textbf{3}};
		\node (n2) at (300: \R)  {\textbf{3}};
		\node (n3) at (360: \R)  {\textbf{6}};
		
		\foreach \from/\to in {n4/n5,n5/n1,n2/n5,n3/n4,n5/n3,n6/n3}
		\draw[thick] (\from) -- (\to);
		\end{tikzpicture}
	}
	\hfill
	\hfill
	\centering
	\subfigure[Relabeled $\mathsf{G}_2$.]{
		\begin{tikzpicture}
		[scale=1.1,every node/.style={draw, circle, thick,minimum size=1.5em,inner sep=1}]
		\node (n6) at (60: \R) {\textbf{5}};
		\node (n4) at (120: \R)  {\textbf{4}};
		\node (n5) at (180: \R)  {\textbf{10}};
		\node (n1) at (240: \R) {\textbf{2}};
		\node (n2) at (300: \R)  {\textbf{1}};
		\node (n3) at (360: \R)  {\textbf{7}};
		
		\foreach \from/\to in {n6/n4,n5/n1,n2/n1,n5/n6,n5/n3,n6/n3}
		\draw[thick] (\from) -- (\to);
		\end{tikzpicture}
	}
	\hspace*{\fill}
	\caption{Relabel graphs. $\Sigma=\{1,2,3,4,5,6,7,8,9,10\}$.}
	\label{fig:relabel}
\end{figure}
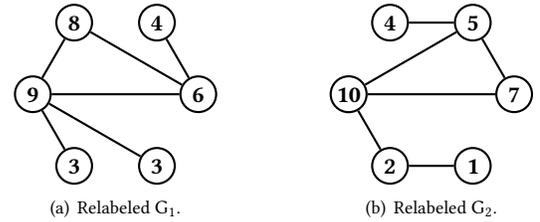

For each truncated BFS tree, we need to hash it to a value which is used for its label. In this paper, we just use the concatenation of the labels of its vertices as a hash method. Note that the child vertices of each parent vertex in the BFS trees are sorted by their label and eigenvector centrality values, from low to high. Thus, each truncated BFS tree is uniquely hashed to a string of vertex labels. For example, in Figure \ref{fig:truncatedtree1}, $\mathsf{T}_1^{(1)}$ can be denoted as $(1,4,1,3,3)$, and $\mathsf{T}_4^{(1)}$ can be denoted as $(3,3,4,2,1,1)$. Now, the label function $l: \mathsf{T}\rightarrow \Sigma$ can assign the same positive integers to the same trees (the same sequences of vertex labels). In our implementation, we use a set to store BFS trees of depth $k$ rooted at each vertex in a dataset of graphs. In this case, the set will only contain unique BFS trees. For BFS trees shown in Figure \ref{fig:truncatedtree1} and Figure \ref{fig:truncatedtree2}, the set will contain $\mathsf{T}_1^{(2)}$, $\mathsf{T}_2^{(2)}$, $\mathsf{T}_1^{(1)}$, $\mathsf{T}_3^{(1)}$, $\mathsf{T}_4^{(2)}$, $\mathsf{T}_5^{(1)}$, $\mathsf{T}_5^{(2)}$, $\mathsf{T}_4^{(1)}$, $\mathsf{T}_6^{(1)}$, and $\mathsf{T}_6^{(2)}$. Note that the truncated BFS trees in the set are sorted lexicographically. We can use the index of each truncated BFS tree in the set as its label. For instance,  $l: \mathsf{T}_1^{(2)}\rightarrow 1$, $l: \mathsf{T}_2^{(2)}\rightarrow 2$, $l: \mathsf{T}_1^{(1)}\rightarrow 3$, $l: \mathsf{T}_3^{(1)}\rightarrow 4$, $l: \mathsf{T}_4^{(2)}\rightarrow 5$, $l: \mathsf{T}_5^{(1)}\rightarrow 6$, $l: \mathsf{T}_5^{(2)}\rightarrow 7$, $l: \mathsf{T}_4^{(1)}\rightarrow 8$, $l: \mathsf{T}_6^{(1)}\rightarrow 9$, and $l: \mathsf{T}_6^{(2)} \rightarrow 10$. If we use the labels of these truncated BFS trees to relabel their root vertices, graphs $\mathsf{G}_1$ and $\mathsf{G}_2$ in Figure \ref{fig:example}(a) and (b) become graphs shown in Figure \ref{fig:relabel}(a) and (b).

One observation is that if two vertices have the same structural identities, their corresponding truncated BFS trees are the same and thus they will have the same new labels. For example, Figure \ref{fig:truncatedtree1}(a) and (b) show two truncated BFS trees on the two vertices with the same label 1 in Figure \ref{fig:example}(a). The two trees are identical, and thus these two vertices' structural identities are identical, and their new labels in Figure \ref{fig:relabel}(a) are also the same. This phenomenon also happens across graphs, e.g., the vertices with label 2 in Figure \ref{fig:example}(a) and (b) also have the same labels in Figure \ref{fig:relabel}(a) and (b) (vertices with the label 4). Figure \ref{fig:truncatedtree1}(d) and (e) show another two truncated BFS trees on the two vertices with label 3 in Figure \ref{fig:example}(a). We can see that they have different structural identities. Thus, by integrating structural identities into path patterns, we can distinguish path patterns at different levels of granularities. If we build truncated BFS trees of depth 0 rooted at each vertex for super paths, the two path patterns $(1,4,3)$ in Figure \ref{fig:example}(a) and $(1,4,3)$ in Figure \ref{fig:example}(b) (The starting vertex is the left-bottom corner vertex with label 1, and the end vertex is the right-most vertex with label 3.) are identical. However, if we build super paths using truncated BFS trees of depth two (as shown in Figure \ref{fig:truncatedtree1} and Figure \ref{fig:truncatedtree2}), the two path patterns become the two new path patterns $(3,9,6)$ and $(2,10,7)$. They are totally different.

\subsection{Tree++}
\begin{definition}[Graph Similarity at the $k$-level of Granularity]
	Given two undirected labeled graphs $\mathsf{G}_1$ and $\mathsf{G}_2$, we build truncated BFS trees of depth $d$ rooted at each vertex in these two graphs. All paths in all of these BFS trees are contained in a set $\mathcal{P}$. For each path in $\mathcal{P}$, we build a truncated BFS tree of depth $k (k\geq 0)$ rooted at each vertex in the path. All super paths are contained in a set $\mathcal{S}^{(k)}$ ($\mathcal{S}^{(k)} = \mathcal{P}$, if $k=0$). The graph similarity at the $k$-level of granularity is defined as follows:
	\begin{equation}
	\mathcal{K}_{pp}^{(k)}(\mathsf{G}_1, \mathsf{G}_2) = \sum_{\mathclap{\substack{S_i^{(k)}\in\mathcal{S}^{(k)}\\
		l(S_i^{(k)})\in\mathcal{U}^{(k)}}}}
	\psi\left( \mathsf{G}_1, l(S_i^{(k)})\right) \psi\left( \mathsf{G}_2, l(S_i^{(k)})\right)
	\end{equation}
	where $\mathcal{U}^{(k)}$ is a set that contains all of unique path patterns at the $k$-level of granularity.
\end{definition}

To make our path pattern graph kernel capture graph similarity at multiple levels of granularities, we formalize the following:
\begin{equation}
\label{equ:tree++}
\mathcal{K}_{Tree++}(\mathsf{G}_1, \mathsf{G}_2) =\sum_{i=0}^k \mathcal{K}_{pp}^{(i)}(\mathsf{G}_1, \mathsf{G}_2)
\end{equation}
We call the above formulation as \textsc{Tree++}. Note that \textsc{Tree++} is positive semi-definite because a sum of positive semi-definite kernels is also positive semi-definite.

In the following, we give the algorithmic details of our \textsc{Tree++} graph kernel in Algorithm 1. Lines 2--8 generate paths for each vertex in each graph. For each vertex $v$, we build a truncated BFS tree of depth $d$ rooted at it. The time complexity of BFS traversal of a graph is $\mathcal{O}(|\mathcal{V}|+|\mathcal{E}|)$, where $|\mathcal{V}|$ is the number of vertices, and $|\mathcal{E}|$ is the number of edges in a graph. For convenience, we assume that $|\mathcal{E}|>|\mathcal{V}|$ and all the $n$ graphs have the same number of vertices and edges. The worst time complexity of our path generation for all the $n$ graphs is $\mathcal{O}\left( n\cdot|\mathcal{V}|\cdot\left(  |\mathcal{E}|+|\mathcal{V}|\right) \right) $. Lines 11--18 generate super paths from paths. The number of paths generated in lines 2--8 is $n\cdot|\mathcal{V}|\cdot\left(  |\mathcal{E}|+|\mathcal{V}|\right) $. For each path, it at most contains $|\mathcal{V}|$ vertices. For each vertex in the path, we need to construct a BFS tree of depth $k$, which costs $\mathcal{O}(|\mathcal{E}|)$. Thus, the worst time complexity of generating super paths for $n$ graphs is $\mathcal{O}\left( n\cdot |\mathcal{V}|^2\cdot |\mathcal{E}|^2+n\cdot |\mathcal{V}|^3\cdot |\mathcal{E}|\right) $. Line 19 sorts the elements in $\mathcal{U}$ lexicographically, of which the time complexity is bounded by $\mathcal{O}(|\mathcal{E}|)$ \cite{shervashidze2011weisfeiler}. Lines 20--21 count the occurrences of each unique path pattern in graphs. For each unique path pattern in $\mathcal{U}^{(i)}$, we need to count its occurrences in each graph. The time complexity for counting is bounded by $\mathcal{O}(q\cdot m)$, where $q$ is the maximum length of all $\mathcal{U}^{(i)} (0\leq i\leq k)$, and $m$ is the maximum length of AllSuperPaths[$j$] ($1\leq j \leq n$). Thus, the time complexity of lines 10--21 is $\mathcal{O}(n\cdot |\mathcal{V}|^2\cdot |\mathcal{E}|^2 +n\cdot |\mathcal{V}|^3\cdot |\mathcal{E}| + n\cdot q\cdot m)$. The time complexity for line 23 is bounded by $\mathcal{O}(n^2\cdot q)$. The worst time complexity of our \textsc{Tree++} graph kernel for $n$ graphs with the depth of $k$ truncated BFS trees for super paths is $\mathcal{O}(k\cdot n\cdot |\mathcal{V}|^2\cdot |\mathcal{E}|^2+k\cdot n\cdot |\mathcal{V}|^3\cdot |\mathcal{E}|+k\cdot n^2\cdot q + k\cdot n\cdot q\cdot m)$.

\begin{algorithm2e}
	\KwIn{A set of graphs $\mathcal{G}=\{\mathsf{G}_1, \mathsf{G}_2, \cdots, \mathsf{G}_n\}$ and their vertex label functions $\mathcal{L}=\{l_1, l_2, \cdots, l_n\}$, $d$, $k$}
	\KwOut{The computed kernel matrix $\matrixSym{K}\in\mathbb{N}^{n\times n}$}
	$\matrixSym{K}\leftarrow$ \upshape zeros(($n$,$n$)), AllPaths$\leftarrow $\{\}\;
	\tcc{Path generation.}
	\For{$i \leftarrow 1$ \KwTo $n$ }{
		Paths$\leftarrow $[]\tcc*[r]{list}
		\ForEach{\upshape vertex $v\in\mathsf{G}_i$ }{
			Build a truncated BFS tree $\mathsf{T}$ of depth $d$ rooted at the vertex $v$\;
			\ForEach{\upshape vertex $v'\in\mathsf{T}$ }{
				Paths.append($P$)\tcc*[r]{$P=(v,v_1,v_2,\cdots,v')$}
			}
		}
		AllPaths[$i$]$\leftarrow$Paths\;
	}
	\tcc{Compute \textsc{Tree++}.}
	\For{$i \leftarrow 0$ \KwTo $k$ }{
		\tcc{Super path generation.}
		$\mathcal{U}^{(i)}\leftarrow $\upshape set( ), AllSuperPaths$\leftarrow $\{\}\;
		\For{$j \leftarrow 1$ \KwTo $n$ }{
			SuperPaths$\leftarrow$ [], Paths$\leftarrow$AllPaths[$j$]\;
			\ForEach{$P\in$ \upshape Paths}{
				\ForEach{\upshape vertex $v$ \upshape in $P$ }{
					Build a truncated BFS tree $\mathsf{T}_{v}$ of depth $i$ rooted at the vertex $v$ in graph $\mathsf{G}_j$\;
				}
				SuperPaths.append($l(S^{(i)})$)\tcc*[r]{$S^{(i)}=(\mathsf{T}_{v_1},\mathsf{T}_{v_2},\cdots), P=(v_1,v_2,\cdots)$}
				$\mathcal{U}^{(i)}$.add($l(S^{(i)})$)\;
			}
			AllSuperPaths[$j$]$\leftarrow$SuperPaths\tcc*[r]{contains all the super paths in graph $\mathsf{G}_j$}
		}
		
		$\mathcal{U}^{(i)}\leftarrow $\upshape sort($\mathcal{U}^{(i)}$)\tcc*[r]{lexicographically}
		
		\For{$j \leftarrow 1$ \KwTo $n$ }{
			\nosemic$\phi(\mathsf{G}_j)\leftarrow \left[ \psi(\mathsf{G}_j,l(S^{(j)}_1)), \psi(\mathsf{G}_j,l(S^{(j)}_2),\cdots,\psi(\mathsf{G}_j,l(S^{(j)}_{|\mathcal{U}^{(i)}|})\right] $\tcc*[r]{count the number $\psi(\mathsf{G}_j,l(S^{(j)}))$ of the occurrences of each path pattern stored in AllSuperPaths[$j$]. $l(S^{(j)}_1), l(S^{(j)}_2),\cdots, l(S^{(j)}_{|\mathcal{U}^{(i)}|}) \in \mathcal{U}^{(i)}$}
		}
		$\matrixSym{\Phi}\leftarrow\left[ \phi(\mathsf{G}_1);\phi(\mathsf{G}_2);\ldots;\phi(\mathsf{G}_n)\right] $\;
		$\matrixSym{K}\leftarrow\matrixSym{K}+\matrixSym{\Phi}\cdot \matrixSym{\Phi}^\intercal$\; 
	}
	
	\Return{$\matrixSym{K}$} \;
	\caption{\textsc{Tree++}}
	\label{alg:tree++}
\end{algorithm2e}

\section{Experimental Setup}
We run all the experiments on a desktop with an Intel Core i7-8700 3.20 GHz CPU, 32 GB memory, and Ubuntu 18.04.1 LTS operating system, Python version 2.7. \textsc{Tree++} is written in Python. We make our code publicly available at Github\footnote{\url{https://github.com/yeweiysh/TreePlusPlus}}.

We compare \textsc{Tree++} with seven state-of-the-art graph kernels, i.e., \textsc{MLG} \cite{kondor2016multiscale}, \textsc{DGK} \cite{yanardag2015deep}, \textsc{RetGK} \cite{zhang2018retgk}, \textsc{Propa} \cite{neumann2016propagation}, \textsc{PM} \cite{nikolentzos2017matching}, \textsc{SP} \cite{borgwardt2005shortest}, and \textsc{WL} \cite{shervashidze2011weisfeiler}. We also compare \textsc{Tree++} with one state-of-the-art graph classification method \textsc{FGSD} \cite{verma2017hunt}  which learns features from graphs and then directly feed them into classifiers. We set the parameters for our \textsc{Tree++} graph kernel as follows: The depth of the truncated BFS tree rooted at each vertex is set as $d=6$, and the depth $k$ of the truncated BFS tree in the super path is chosen from \{0, 1, 2, \ldots, 7\} through cross-validation. The parameters for the comparison methods are set according to their original papers. We use the implementations of \textsc{Propa}, \textsc{PM},  \textsc{SP}, and \textsc{WL} from the GraKeL \cite{siglidis2018grakel} library. The implementations of other methods are obtained from their corresponding websites. A short description for each comparison method is given as follows:
\begin{itemize}
	\item \textsc{MLG} \cite{kondor2016multiscale} is a graph kernel that builds a hierarchy of nested subgraphs to capture graph structures at a range of different scales.
	\item \textsc{DGK} \cite{yanardag2015deep} uses techniques from natural language processing to learn latent representations for substructures extracted by graph kernels such as \textsc{SP} \cite{borgwardt2005shortest}, and \textsc{WL} \cite{shervashidze2011weisfeiler}. Then the similarity matrix between substructures is computed and integrated into the computation of the graph kernel matrices.
	\item \textsc{RetGK} \cite{zhang2018retgk} introduces a structural role descriptor for vertices, i.e., the return probabilities features (RPF) generated by random walks. The RPF is then embedded into the Hilbert space where the corresponding graph kernels are derived.
	\item \textsc{Propa} \cite{neumann2016propagation} leverages early-stage distributions of random walks to capture structural information hidden in vertex neighborhood.
	\item \textsc{PM} \cite{nikolentzos2017matching} embeds graph vertices into vectors and use the Pyramid Match kernel to compute the similarity between the sets of vectors of two graphs.
	\item \textsc{SP} \cite{borgwardt2005shortest} counts the number of pairs of shortest paths which have the same source and sink labels and the same length in two graphs.
	\item \textsc{WL} \cite{shervashidze2011weisfeiler} is based on the Weisfeiler-Lehman test of isomorphism \cite{weisfeiler1968reduction} for graphs. It counts the number of occurrences of each subtrees in graphs.
	\item \textsc{FGSD} \cite{verma2017hunt} discovers family of graph spectral distances and their based graph feature representations to classify graphs.
\end{itemize} 

All graph kernel matrices are normalized according to the method proposed in \cite{feragen2013scalable}. For each entry $\matrixSym{K}(i, j)$, it will be normalized as $\matrixSym{K}(i, j)/\sqrt{\matrixSym{K}(i, i)\matrixSym{K}(j, j)}$. All diagonal entries will be 1. We use 10-fold cross-validation with a binary $C$-SVM \cite{chang2011libsvm} to test classification performance of each graph kernel. The parameter $C$ for each fold is independently tuned from $\left\lbrace1,10,10^{2},10^{3}\right\rbrace $ using the training data from that fold. We repeat the experiments ten times and report the average classification accuracies and standard deviations. We also test the running time of each method on each real-world dataset.

In order to test the efficacy of our graph kernel \textsc{Tree++}, we adopt twelve real-word datasets whose statistics are given in Table \ref{tab:statistics_dataset}. Figure \ref{fig:distributions} shows the distributions of vertex number, edge number and degree in these twelve real-world datasets.

\textbf{Chemical compound datasets}. The chemical compound datasets BZR, BZR\_MD, COX2, COX2\_MD, DHFR, and DHFR\_MD are from the paper \cite{sutherland2003spline}. Chemical compounds or molecules are represented by graphs. Edges represent the chemical bond type, i.e., single, double, triple or aromatic. Vertices represent atoms. Vertex labels represent atom types. BZR is a dataset of 405 ligands for the benzodiazepine receptor. COX2 is a dataset of 467 cyclooxygenase-2 inhibitors. DHFR is a dataset of 756 inhibitors of dihydrofolate reductase. BZR\_MD, COX2\_MD, and DHFR\_MD are derived from BZR, COX2, and DHFR respectively, by removing explicit hydrogen atoms. The chemical compounds in the datasets BZR\_MD, COX2\_MD, and DHFR\_MD are represented as complete graphs, where edges are attributed with distances and labeled with the chemical bond type. NCI1 \cite{wale2008comparison} is a balanced dataset of chemical compounds screened for the ability to suppress the growth of human non-small cell lung cancer.

\textbf{Molecular compound datasets}. The dataset PROTEINS is from \cite{borgwardt2005protein}. Each protein is represented by a graph.
Vertices represent secondary structure elements. Edges represent that two vertices are neighbors along the amino acid sequence or three-nearest neighbors to each other in space. Mutagenicity \cite{riesen2008iam} is a dataset of 4337 molecular compounds which can be divided into two classes mutagen and non-mutagen. The PTC \cite{kriege2012subgraph} dataset consists of compounds labeled according to carcinogenicity on rodents divided into male mice (MM), male rats (MR), female mice (FM) and female rats (FR). 

\textbf{Brain network dataset}. KKI \cite{pan2017task} is a brain network constructed from the whole brain functional resonance image (fMRI) atlas. Each vertex corresponds to a region of interest (ROI), and each edge indicates correlations between two ROIs. KKI is constructed for the task of Attention Deficit Hyperactivity Disorder (ADHD) classification.

\begin{table}[!htb]
	\centering
	\caption{Statistics of the real-world datasets used in the experiments.}
	\label{tab:statistics_dataset}
	\begin{tabular}{l|l|l|l|l|l}
		\toprule
		\multirow{2}{*}{Dataset}         &Size            & Class & Avg.    & Avg.   &Label\\ 
		&                    &\#         &Node\#&Edge\# &\# \\\hline
		BZR                 &405                 &2              &35.75                &38.36           &10         \\ 
		BZR\_MD        &306                &2              &21.30                &225.06         &8            \\ 
		COX2              &467                &2              &41.22                &43.45          &8           \\ 
		COX2\_MD     &303                &2             &26.28                &335.12         &7             \\
		DHFR               &467             &2               &42.43                &44.54         &9           \\ 
		DHFR\_MD      &393             &2               &23.87                &283.01       &7               \\
		NCI1                 &4110           &2              &29.87                  &32.30       &37           \\
		PROTEINS      &1113          &2               &39.06                 &72.82        &3             \\
		Mutagenicity   &4337         &2               &30.32                  &30.77        &14                       \\
		PTC\_MM        &336           &2               &13.97                 &14.32         &20             \\
		PTC\_FR          &351            &2               &14.56                  &15.00        &19               \\
		KKI                    &83             &2               &26.96                &48.42        &190 \\
		\bottomrule
	\end{tabular}
\end{table}

\section{Experimental Results}\label{result}
In this section, we first evaluate \textsc{Tree++} with differing parameters on each real-world dataset, then compare \textsc{Tree++} with eight baselines on classification accuracy and runtime.   
\subsection{Parameter Sensitivity}
In this section, we test the performance of our graph kernel \textsc{Tree++} on each real-world dataset when varying its two parameters, i.e., the depth $k$ of the truncated BFS tree in the super path, and the depth $d$ of the truncated BFS tree rooted at each vertex to extract path patterns. We vary the number of $k$ and $d$ both from zero to seven. When varying the number of $k$, we fix $d=7$. When varying the number of $d$, we fix $k=1$. 

\begin{figure*}[!htb]
	\centering
	\subfigure{\includegraphics [width=0.3\textwidth]{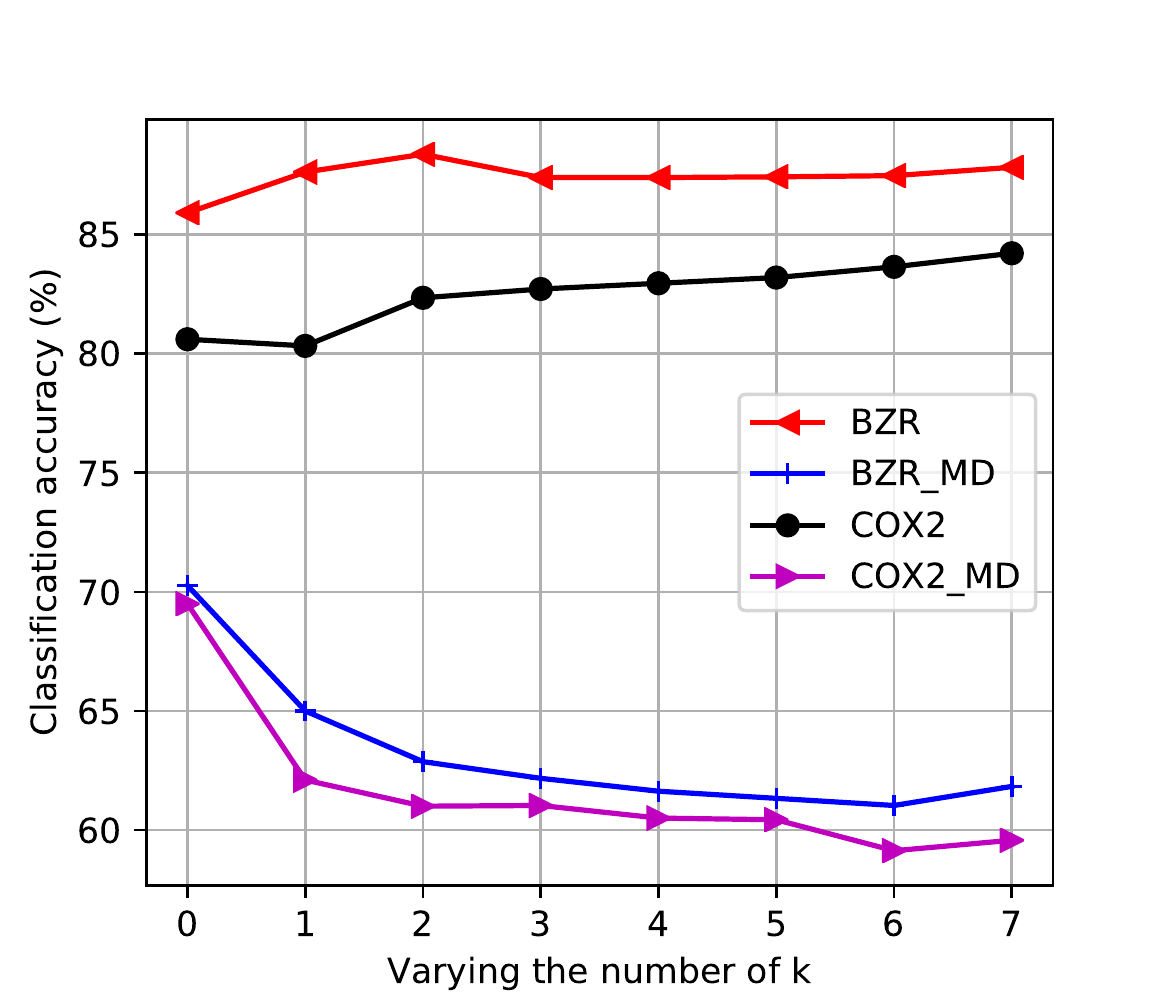}}
	\centering
	\subfigure{\includegraphics [width=0.3\textwidth]{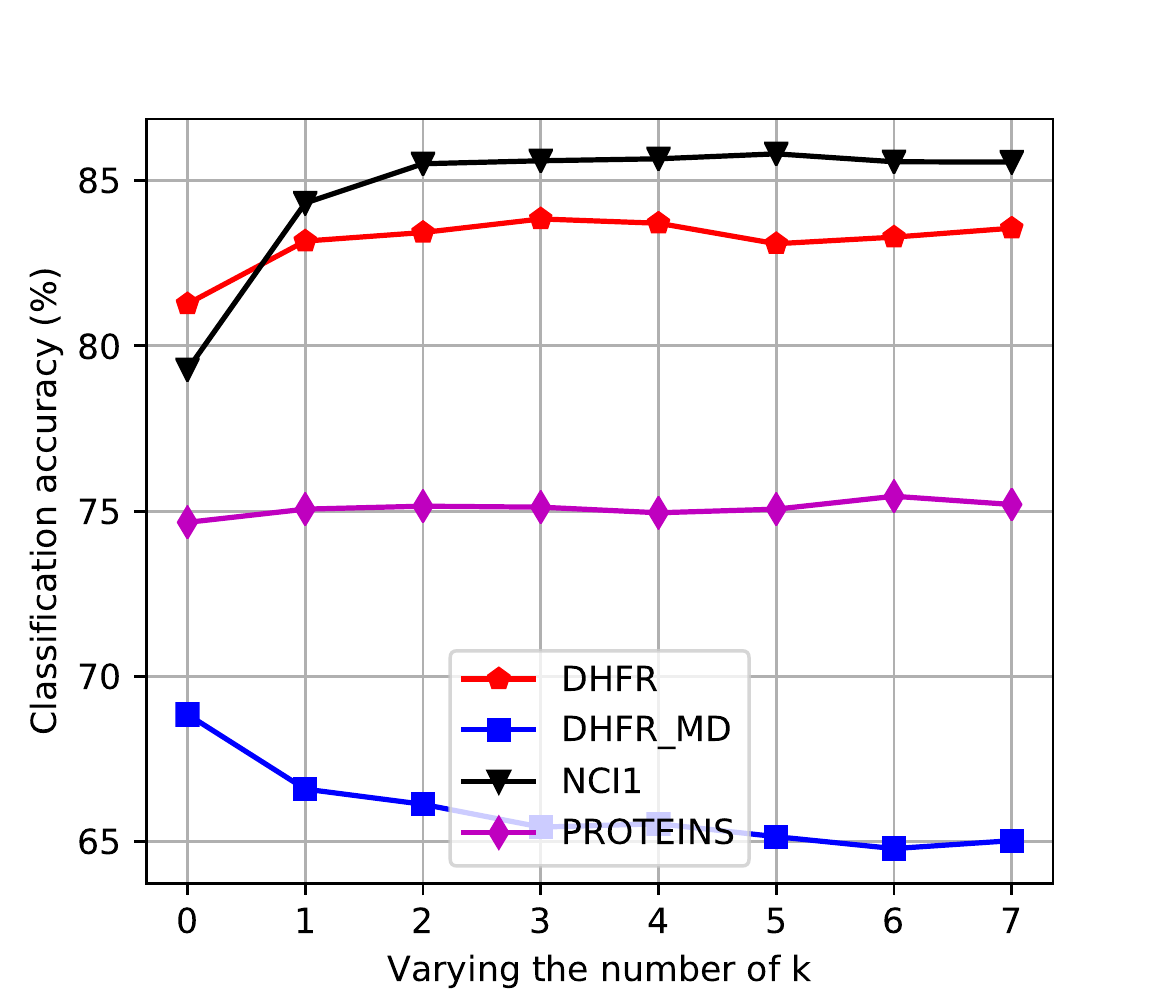}}
	\centering
	\subfigure{\includegraphics [width=0.3\textwidth]{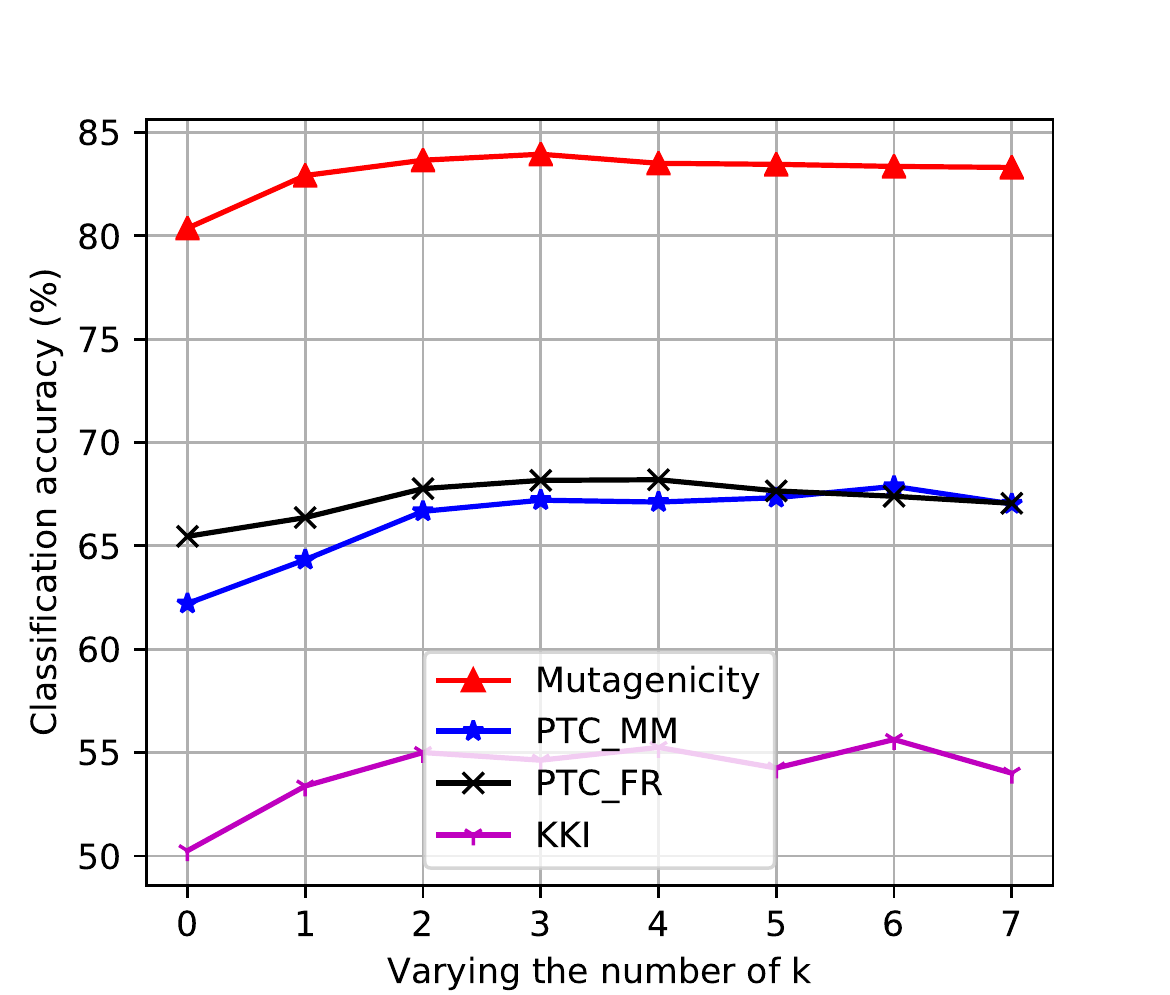}}
	\caption{The classification accuracy of \textsc{Tree++} on each real-world dataset when varying the number of $k$ (the depth of the truncated BFS tree in the super path).}
	\label{fig:varying_k}
\end{figure*}

\begin{figure*}[!htb]
	\centering
	\subfigure{\includegraphics [width=0.6\textwidth]{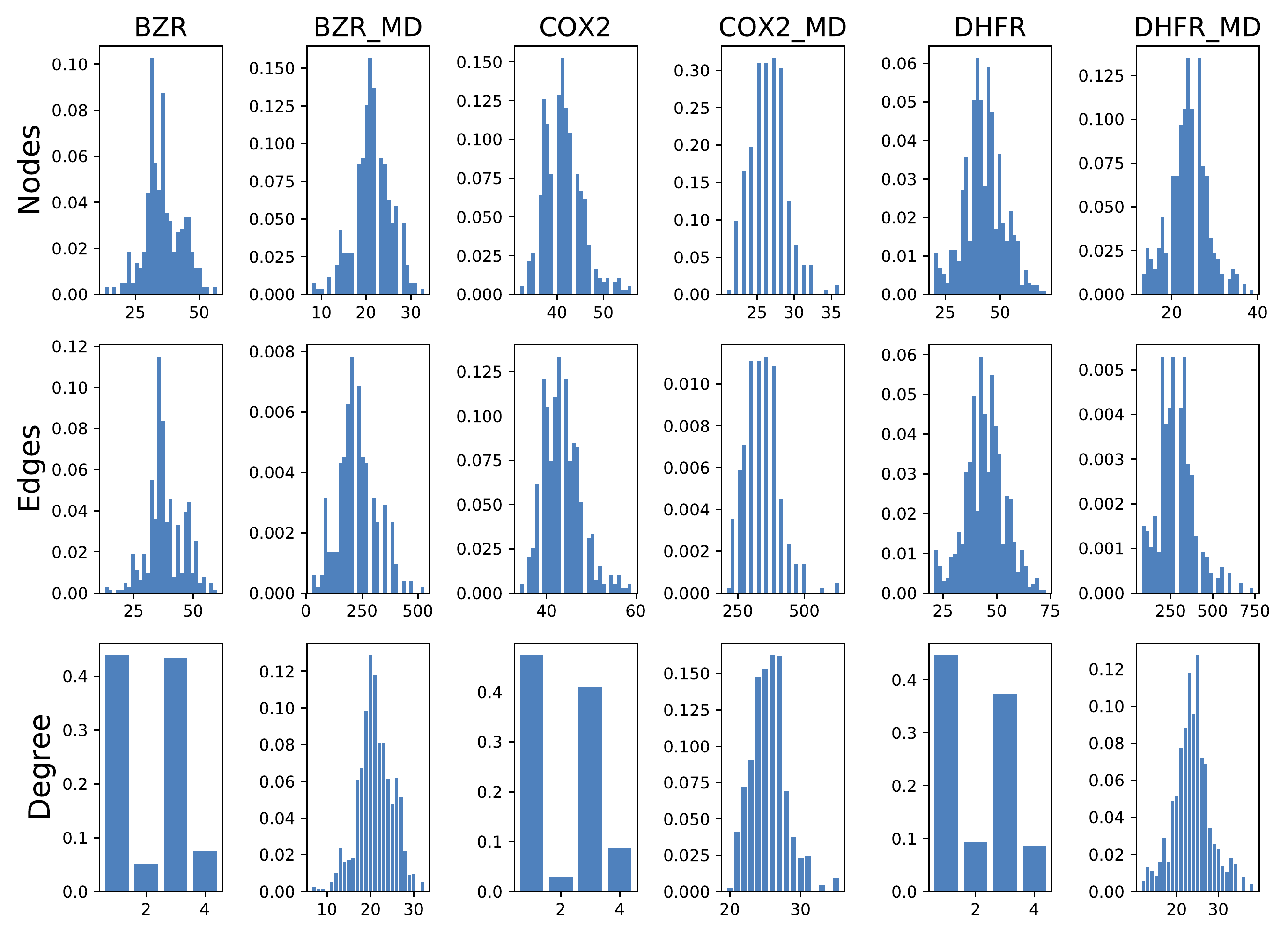}}
	
	\centering
	\subfigure{\includegraphics [width=0.6\textwidth]{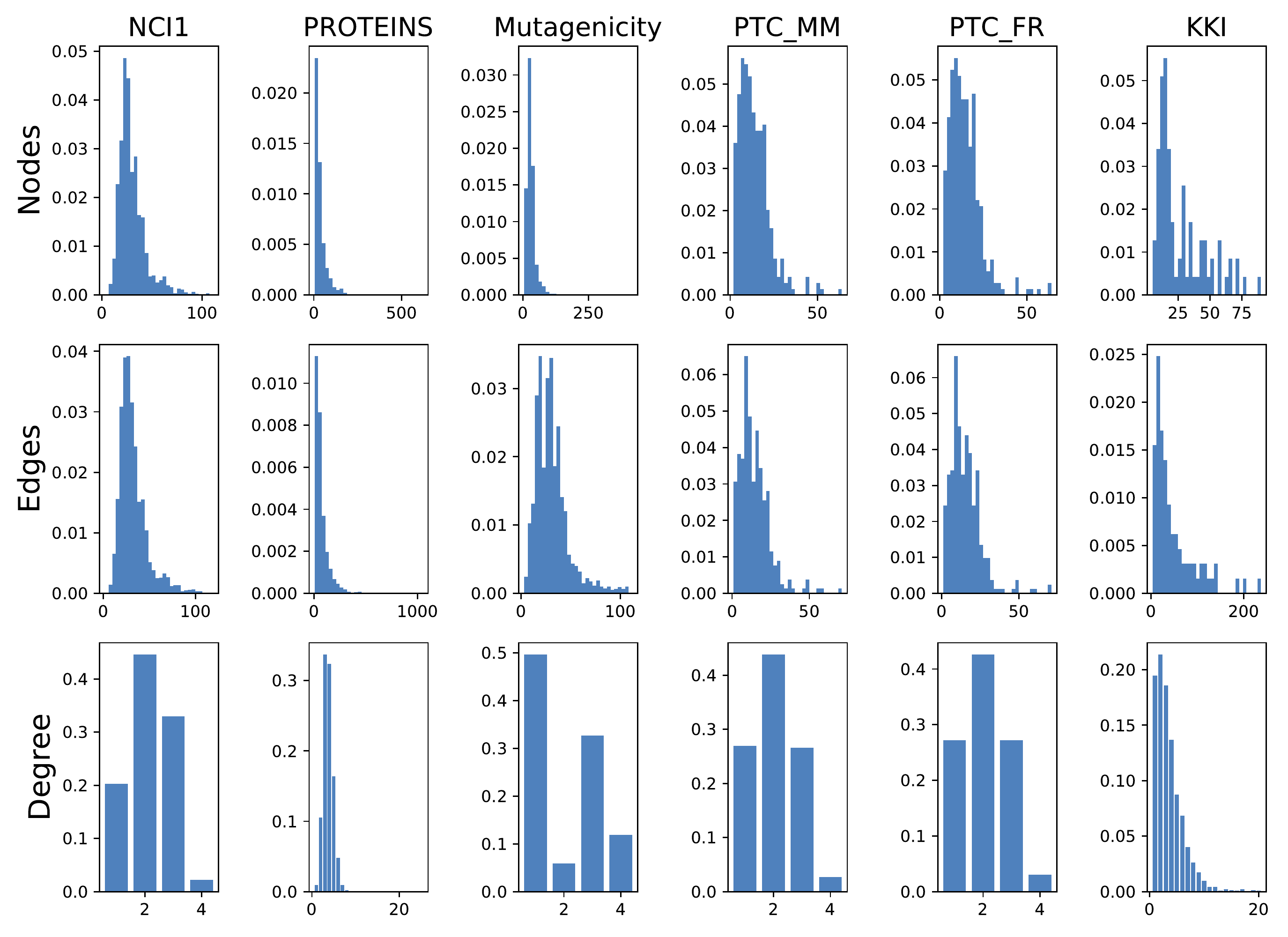}}
	\caption{The rows illustrate the distributions of node number, edge number, and degree in the datasets used in the paper.}
	\label{fig:distributions}
\end{figure*}

Figure \ref{fig:varying_k} shows the classification accuracy of \textsc{Tree++} on each real-world dataset when varying the number of $k$. On the original chemical compound datasets, we can see that \textsc{Tree++} tends to reach better classification accuracy with increasing values of $k$. The tendency is reversed on the derived chemical compound datasets where explicit hydrogen atoms are removed. We can see from Figure \ref{fig:distributions} that compared with the original datasets BZR, COX2, and DHFR, the derived datasets BZR\_MD, COX2\_MD, and DHFR\_MD have more diverse edge and degree distributions, i.e., the edge number and degree vary more than those of the original datasets. In addition, their degree distributions do not follow the power law. For graphs with many high degree vertices, the concatenation of vertex labels as a hash method for BFS trees in super paths can hurt the performance. For example, two BFS trees in super paths may just have one different vertex, which can lead to different hashing. Thus, with increasing values of $k$, two graphs with many high degree vertices tend to be more dissimilar, which is a reason for the decreasing of classification accuracy in datasets BZR\_MD, COX2\_MD, and DHFR\_MD. Since the degree distribution of the brain network dataset KKI follow the power law, we can observe a tendency of \textsc{Tree++} to reach better classification accuracy with increasing values of $k$. On all the molecular compound datasets whose degree distribution also follow the power law, we also observe a tendency of \textsc{Tree++} to reach better classification accuracy with increasing values of $k$. Another observation is that the classification accuracy of \textsc{Tree++} first increase and then remain stable with increasing values of $k$. One explaination is that if smaller values of $k$ can distinguish the structure identities of vertices, larger values of $k$ will not benefit much to the increase of classification accuracy.

\begin{figure*}[!htb]
	\centering
	\subfigure{\includegraphics [width=0.3\textwidth]{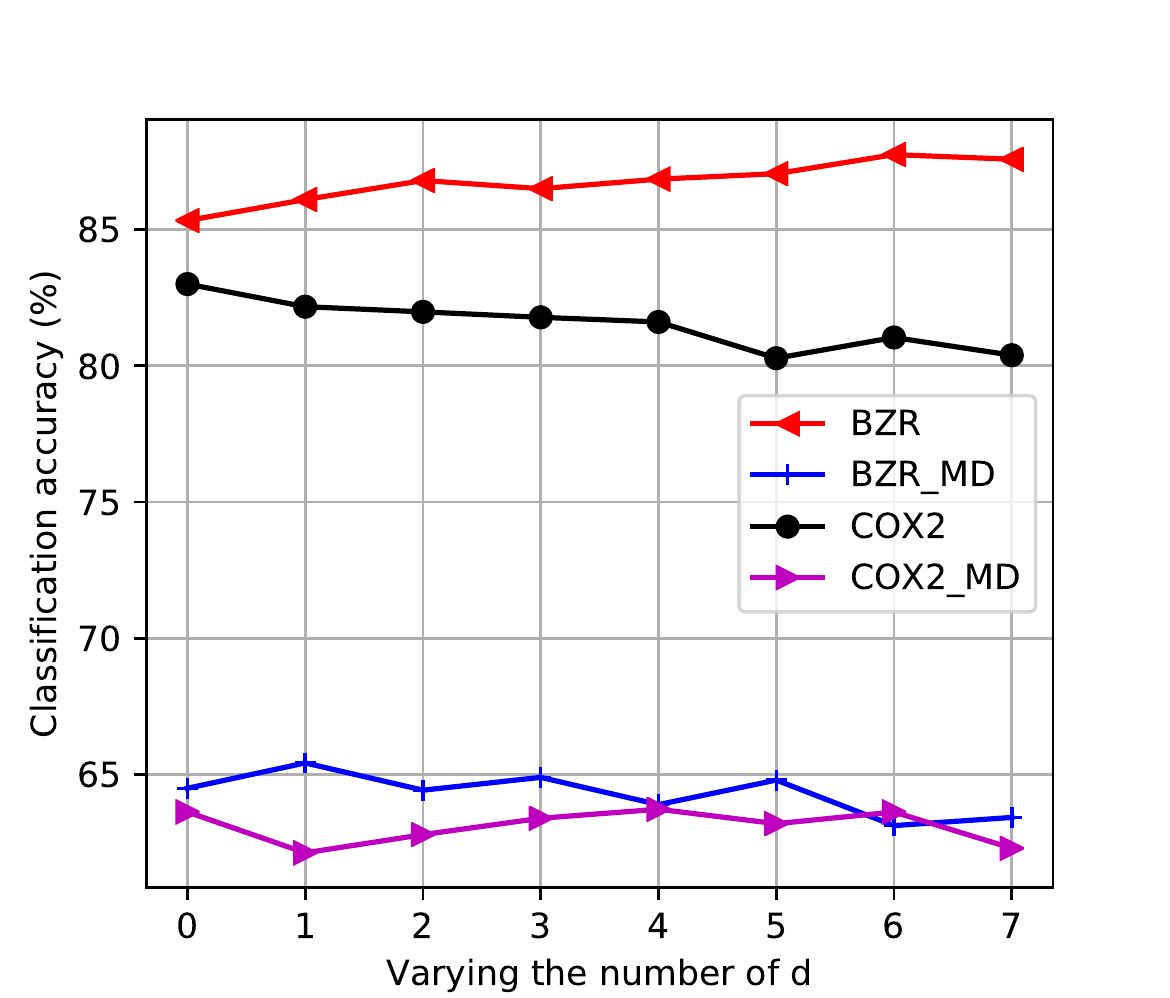}}
	\centering
	\subfigure{\includegraphics [width=0.3\textwidth]{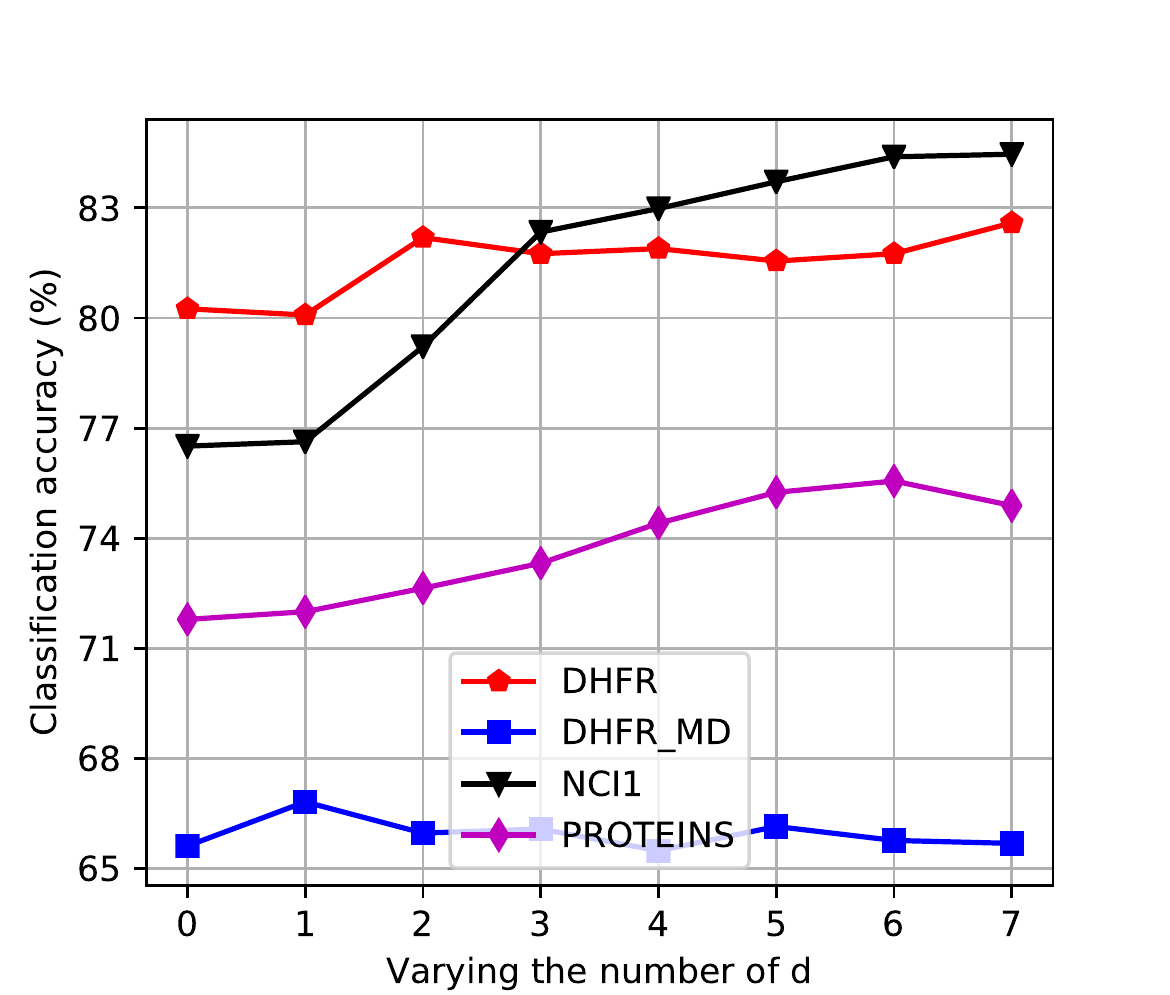}}
	\centering
	\subfigure{\includegraphics [width=0.3\textwidth]{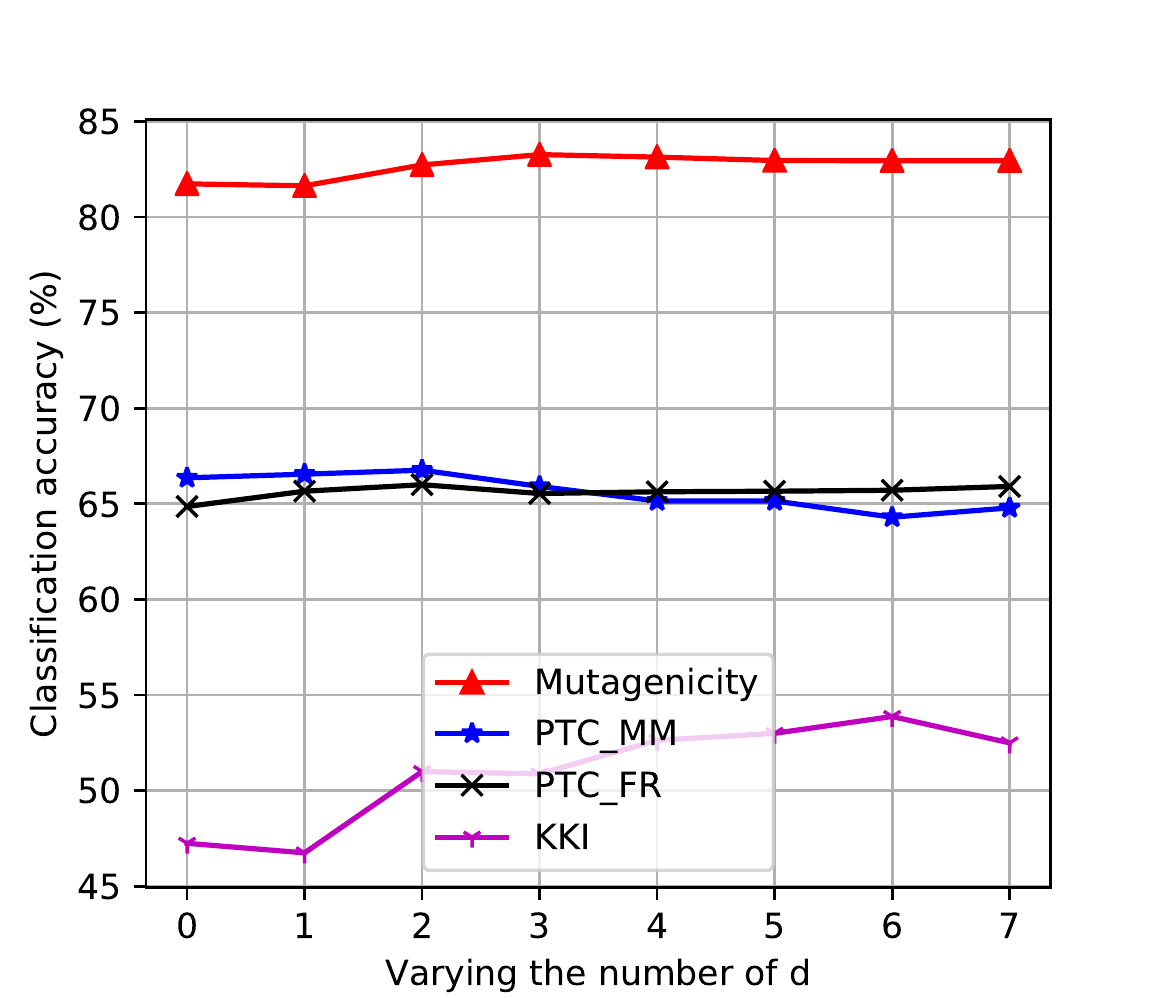}}
	\caption{The classification accuracy of \textsc{Tree++} on each real-world dataset when varying the number of $d$ (the depth of the truncated BFS tree rooted at each vertex to extract path patterns).}
	\label{fig:varying_d}
\end{figure*}

Figure \ref{fig:varying_d} shows the classification accuracy of \textsc{Tree++} on each real-world dataset when varying the number of $d$. On all the chemical compound datasets except COX2, and on all the molecular compound datasets and brain network dataset, \textsc{Tree++} tends to become better with increasing values of $d$. The phenomena are obvious because deep BFS trees can capture more path patterns around a vertex.  

\begin{table*}[!htb]
	\centering
	\caption{Comparison of classification accuracy ($\pm$ standard deviation) of \textsc{Tree++} and its competitors on the real-world datasets.}
	\label{tab:classification}
	\begin{tabular}{l|l|l|l|l|l|l|l|l|l}
		\toprule
		Dataset  &\textsc{Tree++}                      & \textsc{MLG}          & \textsc{DGK}          & \textsc{RetGK}      &\textsc{Propa}            &\textsc{PM}            &\textsc{SP}                &\textsc{WL}           &\textsc{FGSD} \\ \hline
		BZR           &\textbf{87.88}$\pm$1.00     &86.28$\pm$0.59    &83.08$\pm$0.53   &86.30$\pm$0.71     &85.95$\pm$0.85        &82.35$\pm$0.47   &85.65$\pm$1.02      &87.25$\pm$0.77   &85.38$\pm$0.85 \\ 
		BZR\_MD   &\textbf{69.47}$\pm$1.14     &48.87$\pm$2.44    &58.50$\pm$1.52    &62.77$\pm$1.69     &61.53$\pm$1.27         &68.20$\pm$1.24    &68.60$\pm$1.94     &59.67$\pm$1.47   &61.00$\pm$1.35 \\ 
		COX2        &\textbf{84.28}$\pm$0.85     &76.91$\pm$1.14      &78.30$\pm$0.29   &81.85$\pm$0.83     &81.33$\pm$1.36         &77.34$\pm$0.82    &80.87$\pm$1.20     &81.20$\pm$1.05     &78.30$\pm$1.03 \\ 
		COX2\_MD &\textbf{69.20}$\pm$1.69      &48.17$\pm$2.43    &51.57$\pm$1.71      &59.47$\pm$1.66    &55.33$\pm$1.70          &63.60$\pm$0.87   &65.70$\pm$1.66    &56.30$\pm$1.55     &48.97$\pm$1.90 \\
		DHFR         &\textbf{83.68}$\pm$0.59     &79.61$\pm$0.50     &64.13$\pm$0.89    &82.33$\pm$0.66    &80.67$\pm$0.52        &64.59$\pm$1.25    &77.80$\pm$0.98    &82.39$\pm$0.90    &78.13$\pm$0.58 \\ 
		DHFR\_MD &\textbf{68.87}$\pm$0.91     &67.87$\pm$0.12     &67.90$\pm$0.26   &64.44$\pm$0.98     &64.18$\pm$0.97        &66.21$\pm$1.01    &68.00$\pm$0.36     &64.00$\pm$0.47    &66.62$\pm$0.78 \\
		NCI1           &\textbf{85.77}$\pm$0.12       &78.20$\pm$ 0.32   &66.72$\pm$0.29    &84.28$\pm$0.25    &79.71$\pm$0.39         &63.55$\pm$0.44    &73.12$\pm$0.29     &84.79$\pm$0.22     &75.99$\pm$0.51 \\
		PROTEINS  &75.46$\pm$0.47                   &72.01$\pm$0.83     &72.59$\pm$0.51    &75.77$\pm$0.66     &72.71$\pm$0.83         &73.66$\pm$0.67     &\textbf{76.00}$\pm$0.29    &75.32$\pm$0.20      &70.14$\pm$0.67 \\
		Mutagenicity &\textbf{83.64}$\pm$0.27     &76.85$\pm$0.38    &66.80$\pm$0.15    &82.89$\pm$0.18  &81.47$\pm$0.34         &69.06$\pm$0.14    &77.52$\pm$0.13     &83.51$\pm$0.27      &70.71$\pm$0.39 \\
		PTC\_MM    &\textbf{68.03}$\pm$0.61     &61.21$\pm$1.08      &67.09$\pm$0.49   &65.79$\pm$1.76     &64.12$\pm$1.43         &62.27$\pm$1.51      &62.18$\pm$2.22     &67.18$\pm$1.61       &57.88$\pm$1.97 \\
		PTC\_FR     &\textbf{68.71}$\pm$1.29        &64.31$\pm$2.00    &67.66$\pm$0.32    &66.77$\pm$0.99   &65.14$\pm$2.04         &64.86$\pm$0.88    &66.91$\pm$1.46    &66.17$\pm$1.02       &63.80$\pm$1.51 \\
		KKI             &\textbf{55.63}$\pm$1.69      &48.00$\pm$3.64    &51.25$\pm$4.17     &48.50$\pm$2.99    &50.88$\pm$4.17         &52.25$\pm$2.49    &50.13$\pm$3.46    &50.38$\pm$2.77      &49.25$\pm$4.76 \\
		\bottomrule
	\end{tabular}
\end{table*}

\subsection{Classification Results}
Table \ref{tab:classification} shows the classification accuracy of our graph kernel \textsc{Tree++} and its competitors on the twelve real-world datasets. \textsc{Tree++} is superior to all of the competitors on eleven real-world datasets. On the dataset COX2\_MD, the classification accuracy of \textsc{Tree++} has a gain of 5.3\% over that of the second best method \textsc{SP}, and has a gain of 43.7\% over that of the worst method \textsc{MLG}. On the dataset KKI, the classification accuracy of \textsc{Tree++} has a gain of 6.5\% over that of the second best method \textsc{PM}, and has a gain of 15.9\% over that of the worst method \textsc{MLG}. On the datasets DHFR\_MD, Mutagenicity, and PTC\_MM, \textsc{Tree++} is slightly better than \textsc{WL}. On the dataset PROTEINS, \textsc{SP} achieves the best classification accuracy. \textsc{Tree++} achieves the second best classification accuracy. However, the classification accuracy of \textsc{SP} only has a gain of 0.7\% over that of \textsc{Tree++}. To summarize, our \textsc{Tree++} kernel achieves the highest accuracy on eleven datasets and is comparable to \textsc{SP} on the dataset PROTEINS.

\begin{table*}[!htb]
	\centering
	\caption{Comparison of runtime (in seconds) of \textsc{Tree++} and its competitors on the real-world datasets.}
	\label{tab:runtime}
	\begin{tabular}{l|l|l|l|l|l|l|l|l|l}
		\toprule
		Dataset         &\textsc{Tree++}   & \textsc{MLG}  & \textsc{DGK}  & \textsc{RetGK}  &\textsc{Propa}  &\textsc{PM}   &\textsc{SP}   &\textsc{WL} &\textsc{FGSD} \\ \hline
		BZR                 &11.29                      &142.80              &1.60                  &13.70                    &11.76                   &16.80              &12.37             &1.73               &0.73 \\ 
		BZR\_MD        &4.73                        &89.93               &1.23                   &4.22                     &4.89                    &17.15               &17.81              &7.72              &0.07 \\ 
		COX2              &14.57                     &78.29               &2.26                  &15.73                    &7.28                    &6.48                &4.81               &0.92               &0.14 \\ 
		COX2\_MD     &7.83                      &4.42                   &1.10                  &5.62                      &1.71                      &4.67                &2.67               &1.14               &0.07 \\
		DHFR              &26.24                     &200.05               &4.44                &48.95                   &14.17                   &16.01                &12.07             &1.95               &0.22 \\ 
		DHFR\_MD     &8.10                       &19.03                 &1.12                  &10.02                     &3.01                     &6.82                &4.65               &1.49               &0.08 \\
		NCI1               &81.68                    &3315.42              &39.35                &761.45                 &221.84                 &326.43           &22.89             &101.68            &1.39 \\
		PROTEINS&59.56                   &3332.31               &48.83              &49.07                    &27.72                    &32.79              &36.38           &38.66               &0.49 \\
		Mutagenicity &87.09                    &4088.53             &24.67               &735.48                 &526.96                &672.33           &28.03             &94.05              &1.56 \\
		PTC\_MM      &1.99                       &152.69                 &1.08                 &2.84                      &2.44                     &10.00               &1.83               &0.95                 &0.08 \\
		PTC\_FR        &2.19                       &170.05                 &1.14                  &3.16                      &5.73                      &10.01               &2.48               &1.77                 &0.09 \\
		KKI                  &2.00                      &67.58                  &0.65                 &0.40                      &0.57                     &1.25                 &1.27                &0.25                &0.02 \\
		\bottomrule
	\end{tabular}
\end{table*}

\subsection{Runtime}
Table \ref{tab:runtime} demonstrates the running time of every method on the real-world datasets. \textsc{Tree++} scales up easily to graphs with thousands of vertices. On the dataset Mutagenicity, \textsc{Tree++} finishes its computation in about one minute. It costs \textsc{RetGK} about twelve minutes to finish. It even costs \textsc{MLG} about one hour to finish. On the dataset NCI1, \textsc{Tree++} finishes its computation in about one minute, while \textsc{RetGK} uses about twelve minutes and \textsc{MLG} uses about one hour. On the other datasets, \textsc{Tree++} is comparable to \textsc{SP} and \textsc{WL}.

\section{Discussion}
 Differing from the Weisfeiler-Lehman subtree kernel (WL) which uses subtrees (each vertex can appear repeatedly) to extract features from graphs, we use BFS trees to extract features from graphs. In this case, every vertex will appear only once in a BFS tree. Another different aspect is that we count the number of occurrences of each path pattern while WL counts the number of occurrences of each subtree pattern. If the BFS trees used in the construction of path patterns and super paths are of depth zero, \textsc{Tree++} is equivalent to WL using subtree patterns of depth zero; If the BFS trees used in the construction of path patterns are of depth zero, and of super paths are of depth one, \textsc{Tree++} is equivalent to WL using subtree patterns of depth one. In other cases, \textsc{Tree++} and the Weisfeiler-Lehman subtree kernel deviate from each other. \textsc{Tree++} is also related to the shortest-path graph kernel (SP) in the sense that both of them use the shortest paths in graphs. SP counts the number of pairs of shortest paths which have the same source and sink labels and the same length in two graphs. Each shortest-path used in SP is represented as a tuple in the form of  ``(source, sink, length)'' which does not explicitly consider the intermediate vertices. However, \textsc{Tree++} explicitly considers the intermediate vertices. If two shortest-paths with the same source and sink labels and the same length but with different intermediate vertices, SP cannot distinguish them whereas \textsc{Tree++} can. Thus compared with SP, \textsc{Tree++} has higher discrimination power.
 
 As discussed in Section \ref{intro}, WL can only capture the graph similarity at coarse granularities, and SP can only capture the graph similarity at fine granularities. By inheriting merits both from trees and shortest-paths, our method \textsc{Tree++} can capture the graph similarity at multiple levels of granularities. Although MLG can also capture the graph similarity at multiple levels of granularities, it needs to invert the graph Laplacian matrix, which costs a lot of time. \textsc{Tree++} is scalable to large graphs. \textsc{Tree++} is built on the truncated BFS trees rooted at each vertex in a graph. One main problem is that the truncated BFS trees are not unique. To solve this problem, we build BFS trees considering the label and eigenvector centrality values of each vertex. Alternatively, we can also use other centrality metrics such as closeness centrality \cite{sabidussi1966centrality} and betweenness centrality \cite{freeman1977set} to order the vertices in the BFS trees. An interesting research topic in the future is to investigate the effects of using different centrality metrics to construct BFS trees on the performance of \textsc{Tree++}.
 
As stated in Section \ref{relatedWork}, hash functions have been integrated into the design of graph kernels. But they are just adopted for hashing continuous attributes to discrete ones. Conventionally, hash functions are developed for efficient nearest neighbor search in databases. Usually, people first construct a similarity graph from data and then learn a hash function to embed data points into a low-dimensional space where neighbors in the input space are mapped to similar codes \cite{liu2011hashing}. For two graphs, we can first use hash functions such as Spectral Hashing (SH) \cite{weiss2009spectral} or graph embedding methods such as DeepWalk \cite{perozzi2014deepwalk} to embed each vertex in a graph into a vector space. Each graph is represented as a set of vectors. Then, following RetGK \cite{zhang2018retgk}, we can use the Maximum Mean Discrepancy (MMD) \cite{gretton2012kernel} to compute the similairty between two sets of vectors. Finally, we have a kernel matrix for graphs. This research direction is worth exploring in the future. \textsc{Tree++} is designed for graphs with discrete vertex labels. Another research direction in the future is to extend \textsc{Tree++} to graphs with both discrete and continuous attributes.

\section{Conclusion}\label{conclusion}
In this paper, we have presented two novel graph kernels: (1) The path-pattern graph kernel that uses the paths from the root to every other vertex in a truncated BFS tree as features to represent a graph; (2) The \textsc{Tree++} graph kernel that incorporates a new concept of super path into the path-pattern graph kernel and can compare the graph similarity at multiple levels of granularities. \textsc{Tree++} can capture topological relations between not only individual vertices, but also subgraphs, by adjusting the depths of truncated BFS trees in the super paths. Empirical studies demonstrate that \textsc{Tree++} is superior to other well-known graph kernels in the literature regarding classification accuracy and runtime.

\section*{Acknowledgment}
The authors would like to thank anonymous reviewers for their constructive and helpful comments. This work was supported partially by the National Science Foundation (grant \# IIS-1817046) and by the U. S. Army Research Laboratory and the U. S. Army Research Office (grant \# W911NF-15-1-0577).

\bibliographystyle{ACM-Reference-Format}
\bibliography{references}

\end{document}